\documentclass{article} % For LaTeX2e
\usepackage{iclr2026_conference,times}

\usepackage{amsmath,amsfonts,bm}

\def\1{\bm{1}}

\DeclareMathAlphabet{\mathsfit}{\encodingdefault}{\sfdefault}{m}{sl}
\SetMathAlphabet{\mathsfit}{bold}{\encodingdefault}{\sfdefault}{bx}{n}

\usepackage{iclr2026_conference,times}
\usepackage[utf8]{inputenc} % allow utf-8 input
\usepackage[T1]{fontenc}    % use 8-bit T1 fonts
\usepackage{hyperref}       % hyperlinks
\usepackage{url}            % simple URL typesetting
\usepackage{booktabs}       % professional-quality tables
\usepackage{amsfonts}       % blackboard math symbols
\usepackage{nicefrac}       % compact symbols for 1/2, etc.
\usepackage{microtype}      % microtypography
\usepackage{xcolor}         % colors
\usepackage{color, colortbl}
\usepackage{bm}
\usepackage{amsmath} 
\usepackage{mathtools}
\usepackage{enumitem}
\usepackage{amssymb}
\usepackage{wrapfig}
\usepackage{algorithm}
\usepackage{algpseudocode}
\usepackage{amsthm}

\usepackage[normalem]{ulem}

\usepackage{caption}
\usepackage{subcaption}
\usepackage[export]{adjustbox}
\usepackage{multirow}

\def\eg{\emph{e.g.}} 

\def\ie{\emph{i.e.}}

\newcommand{\tableCellHeight}{1}
\newcommand{\tabstyle}[1]{
  \setlength{\tabcolsep}{#1}
  \renewcommand{\arraystretch}{\tableCellHeight}
  \centering
  \small
}

\definecolor{tabhighlight}{HTML}{e5e5e5}
\definecolor{white}{rgb}{1,1,1}
\definecolor{lightCyan}{rgb}{0.925,1,1}

\newcolumntype{b}{>{\columncolor{lightCyan}}c}

\newcommand{\ours}{\texttt{CoDi}}
\newcommand{\oursIT}{\texttt{IT}}
\newcommand{\oursIR}{\texttt{IR}}

\renewcommand{\cite}{\citep}

\title{CoDi: Subject-Consistent and Pose-Diverse \\
Text-to-Image Generation}

\author{Zhanxin Gao\textsuperscript{1} \quad Beier Zhu\textsuperscript{2} \quad Liang Yao\textsuperscript{3} \quad Jian Yang\textsuperscript{1} \quad Ying Tai\textsuperscript{1}\thanks{indicates corresponding author.} 
\\
$^1$Nanjing University, $^2$Nanyang Technological University  $^3$Vipshop \\
{\texttt{zxgao@smail.nju.edu.cn} \quad 
\texttt{yingtai@nju.edu.cn}}
}

\iclrfinalcopy % Uncomment for camera-ready version, but NOT for submission.
\begin{document}

\maketitle

\begin{abstract}
 Subject-consistent generation (SCG)—aiming to maintain a consistent subject identity across diverse scenes—remains a  challenge for text-to-image (T2I) models.
Existing training-free SCG methods often achieve consistency at the cost of layout and pose diversity, hindering expressive visual storytelling. 
To address the limitation, we propose subject-\texttt{Co}nsistent and pose-\texttt{Di}verse T2I framework, dubbed as \ours, that enables consistent subject generation with diverse pose and layout.   
Motivated by the progressive nature of diffusion, where coarse structures emerge early and fine details are refined later, \ours~adopts a two-stage strategy: \texttt{I}dentity \texttt{T}ransport (\oursIT) and \texttt{I}dentity \texttt{R}efinement (\oursIR).  
  \oursIT~operates in the early denoising steps, using optimal transport to transfer identity features to each target image in a pose-aware manner. This promotes subject consistency while preserving pose diversity.
   \oursIR~is applied in the later denoising steps, selecting the most salient identity features to further refine subject details.
Extensive qualitative and quantitative results on subject consistency, pose diversity, and prompt fidelity demonstrate that \ours~achieves both better visual perception and stronger performance across all metrics. The code is provided in \url{https://github.com/NJU-PCALab/CoDi}.

\end{abstract}
% \vspace{-0.4cm}
\section{Introduction}
% \vspace{-0.3cm}
While text-to-image (T2I)~\cite{ramesh2022hierarchical,saharia2022photorealistic,rombach2022high,blattmann2023align} models excel in high-quality image generation~\cite{rombach2022high,mou2024t2i}, they struggle to maintain subject consistency across multiple scenes. 
Subject-consistent generation (SCG) aims to synthesize images of the same subject across diverse contextual prompts with three key objectives: (1) ensuring subject consistency across generated instances, (2) promoting layout and pose diversity across different instances to avoid repetitive or overly similar compositions, and (3) maintaining prompt fidelity to accurately reflect the semantics of each prompt. 
The capability enables numerous practical applications including multi-scene narrative for visual storytelling, customizable character design for animation and gaming, and coherent illustration sequences for graphic novels. 

\begin{figure}[t]
  \centering
  \includegraphics[width=1.0\textwidth]{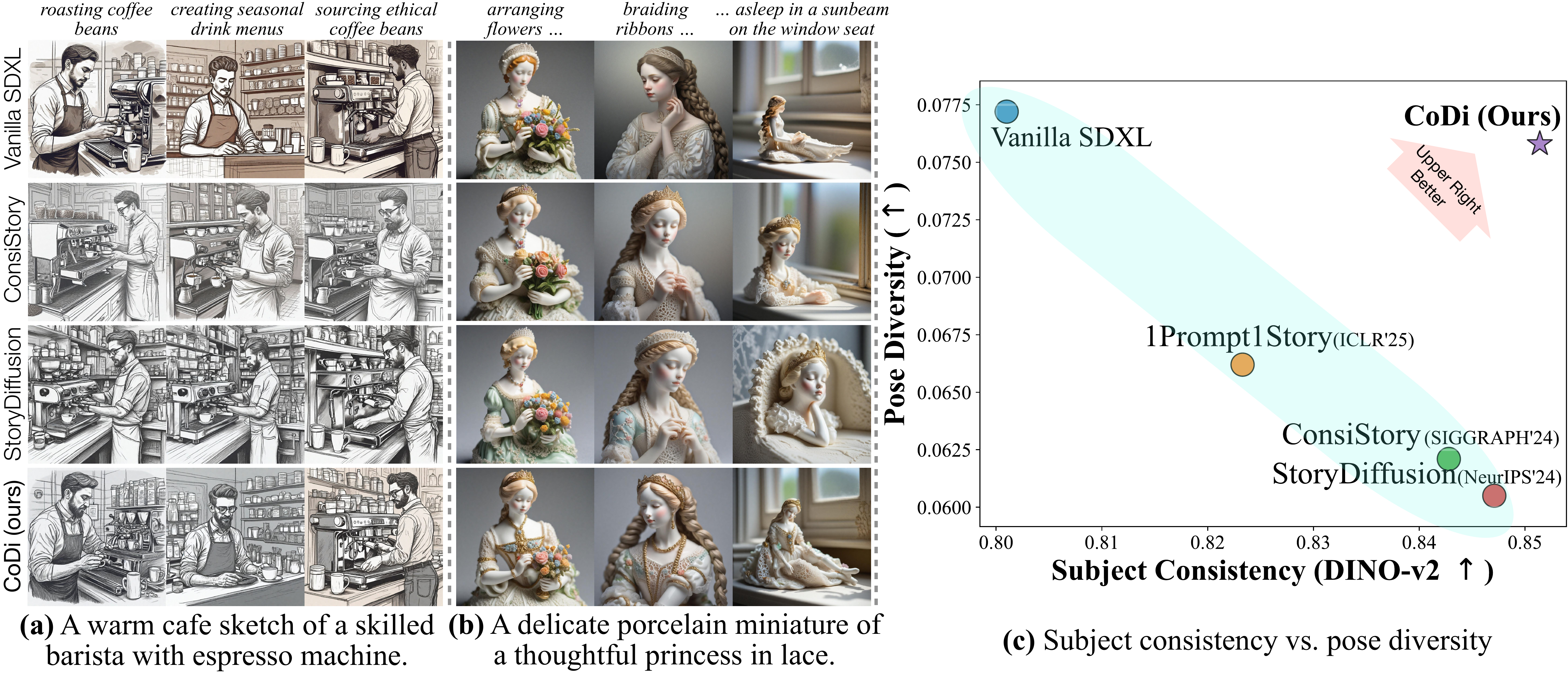}
  \caption{Comparison of  subject-consistent generation methods: Vanilla SDXL~\cite{podell2023sdxl}, ConsiStory~\cite{tewel2024training}, StoryDiffusion~\cite{zhou2024storydiffusion} and \ours~(ours). \textbf{(a\&b)} Existing methods sacrifice pose diversity for subject consistency, \eg, ConsiStory produces similar poses in Figure~\ref{fig:teaser}~(a); and the lower right with hands placed in front in Figure~\ref{fig:teaser}~(b). In contrast, \ours~generates consistent subjects, while matching the pose diversity of Vanilla SDXL.
 % \textbf{(c)} We report two metrics to assess pose quality: 1) fidelity, measuring the distance to the pose of SDXL, and 2) diversity (see Sec.\ref{sec:expsetup} for details). Our \ours~attains the best performance on both metrics. 
 \textbf{(c)} Subject consistency vs. pose diversity. Current methods struggle to balance the two, whereas \ours~achieves both effectively.  
 }
  \label{fig:teaser}
\end{figure}

Current SCG methods~\cite{kopiczko2023vera,ye2023ip} primarily rely on training-intensive optimization~\cite{avrahami2024chosen} or mapping networks~\cite{ruiz2024hyperdreambooth,gal2023designing} to bind subjects to latent representations. These approaches often require computationally expensive fine-tuning per subject or depend on domain-specific encoders, limiting scalability and generalizability. 
Training-free methods~\cite{tewel2024training,zhou2024storydiffusion} have gained significant attention 
due to their elimination of parameter tuning, strong generalization capabilities, and broad compatibility with diverse diffusion architectures.
Current training-free methods—ConsiStory~\cite{tewel2024training} and StoryDiffusion~\cite{zhou2024storydiffusion}—enhance subject consistency by sharing self-attention keys and values across generated images.
However, as noted in their limitations~\cite{tewel2024training,Hertz_2024_CVPR} and evident in Fig.~\ref{fig:teaser}, these methods often achieve high consistency at the cost of severely reduced layout and pose diversity, making it challenging to balance all three objectives.

To better balance the three objectives, we propose a training-free framework—subject-\texttt{Co}nsistent and pose-\texttt{Di}verse generation, dubbed \ours—that achieves strong subject consistency while preserving diverse poses. 
Motivated by the progressive nature of diffusion models~\cite{yue2024exploring}—which shows that low-frequency attributes like pose and layout are formed in early denoising steps, while high-frequency details such as facial features emerge later—our \ours~adopts a two-stage strategy: Identity Transport (\oursIT) and Identity Refinement (\oursIR).
During the early denoising steps, \oursIT~uses optimal transport to align each target image’s features with the reference identity features. Intuitively, this resembles mosaicking: assembling the subject using visual pieces from the reference image, rearranged to match the target pose—thus naturally preserving identity and keeping the original pose. In the later denoising steps, \oursIR~further refines subject consistency by guiding each target image to attend to the most salient identity attributes via cross-attention.
% As shown in Figure~\ref{fig:teaser}, \ours~achieves superior visual results in both subject consistency and pose diversity. We further evaluate pose quality using two metrics: (1) pose fidelity, measuring the distance to the pose of Vanilla SDXL, and (2) pose diversity, quantifying pose variance. \ours~achieves the best performance on both.
As shown in Figure~\ref{fig:teaser}, \ours~achieves superior visual results in both subject consistency and pose diversity, and quantitatively demonstrates advantages over existing methods in balancing this trade-off.

% We evaluate our method on the existing subject-consistent T2I generation benchmark ConsiStory+~\cite{liu2025one}. 
We evaluate our method on the existing T2I SCG benchmark ConsiStory+~\cite{liu2025one}. 
Compared to other training-free approaches, both quantitative and qualitative results validate that our framework achieves better subject consistency while preserving richer layout and pose diversity. 
It demonstrates a superior trade-off among subject consistency, pose diversity, and prompt fidelity.
 Further analysis is also provided to demonstrate \ours's advantages in pose diversity.

% \vspace{-0.2cm}
\section{Related Work}
% \vspace{-0.3cm}
To steer T2I generation with diffusion models~\cite{rombach2022high,podell2023sdxl,nan2024openvid,esser2024scaling,Zhu_2025_ICCV,wang2025paralleldiffusionsolverresidual,hu2023transformer,lin2025pancap,zhou2025explore,lin2025coopdiff,wang2025adaptive}, various methods have been proposed to incorporate control signals such as depth maps, edge maps, and segmentation~\cite{mei2025power,zhang2023adding,yang2023reco,Lei_2025_CVPR,chen2025detailtrainingfreeenhancertexttoimage}. Among them, subject consistency (a.k.a identity preservation) has attracted growing attention, aiming to generate a set of images conditioned on a specified subject.
Existing subject-consistent generation (SCG) methods can be broadly categorized into two groups: training-based and training-free.

\noindent\textbf{Training-based SCG.} 
Training-based methods require either (1) fine-tuning on additional training data~\cite{yang2024seed,li2024photomaker,li2019storygan,betker2023improving,liu2024intelligent} or (2) test-time optimization using reference images~\cite{roich2022pivotal,gal2023designing,kumari2023multi,xiao2024fastcomposer}. The first line of work, represented by StoryDALL-E~\cite{maharana2022storydall} and Make-A-Story~\cite{rahman2023make}, incorporates additional modules to capture subject information, followed by fine-tuning on large datasets to enable direct control over the subject given a reference image. The second line of work, exemplified by DreamBooth~\cite{ruiz2023dreambooth} and Textual Inversion~\cite{gal2022image}, optimizes model parameters or token embeddings on the given test images to inject subject identity. Despite their success in maintaining subject consistency, training-based methods suffer from high training costs or significant test-time latency. In contrast, our \ours~is training-free and introduces only mild additional latency.
% Despite their success in maintaining subject consistency, training-based methods suffer from high training costs or significant test-time latency. In contrast, our \ours~is training-free and introduces only mild additional latency.

\noindent\textbf{Training-free SCG.} 
Training-free methods circumvent the need for iterative tuning of model parameters.
For instance, 1Prompt1Story~\cite{liu2025one} improves consistency by aligning prompt embeddings across generations. However, textual embedding control alone is insufficient to to enforce consistency, often resulting in subject drift. 
% The current leading methods, ConsiStory~\cite{tewel2024training} and StoryDiffusion~\cite{zhou2024storydiffusion}, adopt {attention-based mechanisms} to promote subject consistency by sharing self-attention keys and values across generated images.
The current leading methods, ConsiStory~\cite{tewel2024training} and StoryDiffusion~\cite{zhou2024storydiffusion}, adopt attention-based mechanisms to promote subject consistency by sharing self-attention keys and values.
However, as noted in their limitation discussions~\cite{tewel2024training,Hertz_2024_CVPR}, \textit{applying attention across a set of images reduces pose diversity}.  
To address this issue, our \ours~explicitly preserves diversity and promotes consistency by aligning early-stage features between the target and reference images via optimal transport.

\section{Method}
Our $\ours$ consists of two stages: Identity Transport (\oursIT) and Identity Refinement (\oursIR). Our \oursIT~operates in the early denoising stage to transport identity features from the reference image while preserving the pose and background of the target images. \oursIR~is applied in later denoising stages to refine subject consistency in fine-grained details. This two-stage design is inspired by~\cite{yue2024exploring}, which reveals that low-frequency attributes such as pose and layout are determined early in the denoising timesteps, whereas high-frequency components like facial details emerge in later steps.
We begin with the setup of subject-consistent generation (SCG), a review of attention-based SCG methods and a brief introduction of optimal transport.

\subsection{Preliminaries}
\begin{figure}[t]
  \centering
  \includegraphics[width=1.0\linewidth]{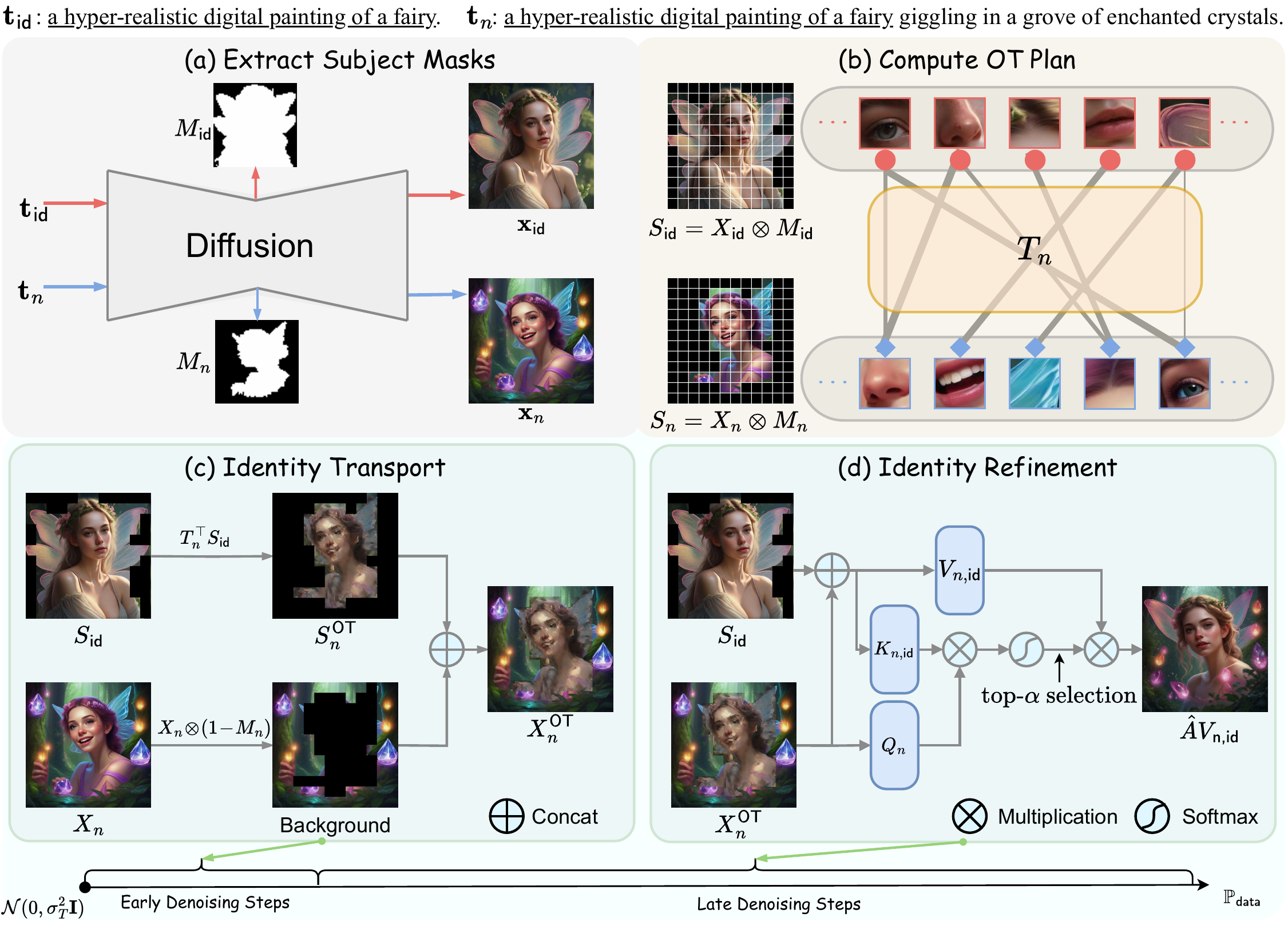}
  \caption{Illustration of our \ours. 
  \textbf{(a)} Extract subject masks ($M_\mathsf{id}$ and $M_n$) by averaging the image-text cross-attention at the final denoising timestep for subject-related tokens (\eg, \texttt{``fairy''}).
  \textbf{(b)} Compute the OT plan $T_n$ using the cost matrix $C$ and the probability masses $\mathbf{a}$ and $\mathbf{b}$ (detailed in Sec.~\ref{section:stage1}). 
  \textbf{(c)} Identity transport (\oursIT) operates in the early denoising steps to 
  transfer reference subject features to targe images in a pose-aware manner. 
  \textbf{(d)} Identity refinement (\oursIR) operates in the late denoising steps to refine subject details using selective cross-image attention mechanism.}
  \label{fig:method}
\end{figure}

\noindent\textbf{Setup.} SCG aims to synthesize a batch of images that share the same subject identity across diverse scenes. Formally, given a set of $N$ textual prompts $\{\mathbf{t}_n\}_{n=1}^N$, where each prompt is composed of a shared identity prompt  $\mathbf{t}_\mathsf{id}$ and a unique attribute prompt $\mathbf{a}_n$, \ie, $\mathbf{t}_n = [\mathbf{t}_\mathsf{id}, \mathbf{a}_n]$. 
For instance, given $\mathbf{t}_1=$\texttt{``\uline{A hyper-realistic digital painting of a fairy} giggling in a grove of enchanted crystals''} and $\mathbf{t}_2=$\texttt{``\uline{A hyper-realistic digital painting of a fairy} lost in a maze of giant sunflowers''},  the identity prompt is $\mathbf{t}_\mathsf{id} =$\texttt{``{a hyper-realistic digital painting of a fairy}''}, and the attribute prompts are $\mathbf{a}_1=$ \texttt{``giggling in a grove of enchanted crystals''} and $\mathbf{a}_2=$ \texttt{``lost in a maze of giant sunflowers''}. 
We refer to the image generated from the identity prompt $\mathbf{t}_\mathsf{id}$ as the \textit{reference image}, denoted as $\mathbf{x}_\mathsf{id}$.
The objective is to generate \textit{target images} $\{\mathbf{x}_n\}_{n=1}^N$ that depict a visually consistent subject with $\mathbf{x}_\mathsf{id}$, while capturing the scene-specific attributes described in ${\mathbf{a}_n}$. See Figure~\ref{fig:method} for a concrete example.

\noindent\textbf{Review of cross-image attention SCG.}
The current leading training-free SCG methods, ConsiStory~\cite{tewel2024training} and StoryDiffusion~\cite{zhou2024storydiffusion}, adopt attention-based strategies that extend the standard self-attention to cross-image attention mechanism. 
Formally, let $\{X_n\}_{n=1}^N$ denote the features of the target images $\{\mathbf{x}_n\}_{n=1}^N$. For generating $i$-th image, standard self-attention first projects $X_i$ to queries $Q_i$, keys $K_i$, and values $V_i$, then compute 
\begin{equation}
    Z_i=\mathrm{Attn}(Q_i,K_i,V_i)=\mathrm{softmax}\left(\frac{Q_iK_i^\top}{\sqrt{d}}\right)V_i,
\end{equation}
where $d$ is the feature dimension. Let $\oplus$ denote matrix concatenation. We compute the concatenated keys and values as $K_{1:N}=[K_1\oplus ...\oplus K_N]$ and $V_{1:N}=[V_1\oplus ...\oplus V_N]$, respectively. To enhance consistency, cross-image attention mechanism allows the feature of the $i$-th image, $X_i$, to attend to the values $V_{1:N}$ of other images using their corresponding keys $K_{1:N}$.
\begin{equation}
    Z_i=\mathrm{Attn}(Q_i,K_{1:N},V_{1:N})=\mathrm{softmax}\left(\frac{Q_iK_{1:N}^\top}{\sqrt{d}}\right)V_{1:N}.
\end{equation}
While both SCG methods adopt cross-image attention, they differ slightly in implementation: ConsiStory~\cite{tewel2024training} limits attention to masked subject regions, whereas StoryDiffusion~\cite{zhou2024storydiffusion} randomly samples tokens from all regions without subject constraints. 

As discussed in their limitations~\cite{tewel2024training,Hertz_2024_CVPR}, attention-based methods significantly reduce layout and pose diversity. We conjecture that attending to a shared pool of keys and values \textit{entangles feature updates across images, implicitly aligning spatial layouts and poses.} To mitigate this, prior work~\cite{tewel2024training} introduces components such as attention dropout and query blending.
However, these additions increase computational overhead and still fail to recover pose diversity (as shown in Figure~\ref{fig:teaser}).
% However, these additions increase computational overhead and still fail to recover pose diversity (as shown in Figure~\ref{fig:teaser}).
In this paper, we draw inspiration from structural learning to simultaneously preserve subject consistency and pose diversity by transporting identity features to each target image via optimal transport.

\noindent\textbf{Optimal transport.} Optimal transport (OT)~\cite{villani2008optimal,monge1781memoire,zhu2025dynamicmultimodalprototypelearning} provides a framework for measuring the distance between two distributions. Specifically, given two sets of support features $\{\mathbf{v}_m\}_{m=1}^M$ and $\{\mathbf{u}_n\}_{n=1}^N$, we define two discrete distributions $\mathbb{P}$ and $\mathbb{Q}$ as:\footnote{We slightly abuse the notations $\mathbf{x}$ and $N$, which here do not refer to an image or the number of target images.} 
\begin{equation}
    \mathbb{P}(\mathbf{x})=\sum_{m=1}^{M}a_m\delta(\mathbf{v}_m-\mathbf{x}), \quad \mathbb{Q}(\mathbf{x})=\sum_{n=1}^{N}b_n\delta(\mathbf{u}_n-\mathbf{x})
\end{equation}
where $\delta(\cdot)$ denotes the Dirac function, and ${a_m}$, ${b_n}$ denote the associated probabilities that sum to 1, respectively. Let $\mathbf{a}=[a_1,...,a_M]^\top$ and $\mathbf{b}=[b_1,...,b_N]^\top$. Given a cost matrix $C \in \mathbb{R}^{M \times N}$, where each entry $C(m, n)$ denotes the transport cost between $\mathbf{v}_m$ and $\mathbf{u}_n$ (typically defined by their similarity), the OT distance between $\mathbb{P}$ and  $\mathbb{Q}$ is defined as:
\begin{equation}\label{eq:ot}
    d_\mathsf{OT}(\mathbb{P},\mathbb{Q};C)=\underset{T\geq0}{\min}\langle T,C\rangle ,\ \mathrm{s.t.}\ T\mathbf{1}_M=\mathbf{a}, T^\top \mathbf{1}_N=\mathbf{b}, 
\end{equation}
where $T \in \mathbb{R}^{M\times N}$ is the transport plan, with $T(m, n) \geq 0$ representing the amount of mass moved from $\mathbf{v}_m$  to  $\mathbf{u}_n$, $\langle,\rangle$ denotes the Frobenius inner product, $\mathbf{1}_M$ is $M$-dimensional all-one vector. 

\subsection{Identity Transport}
\label{section:stage1}
Our \oursIT~operates in the early denoising steps (\eg, the first 10 of 50 total steps) to independently transport identity features from the reference image $\mathbf{x}_\mathsf{id}$ to each target image $\mathbf{x}_n $ for all $ n \in [1,N]$.
Our \oursIT~begins by extracting subject features from masked  regions.

\textbf{Extract subject features.} Masking out background regions offers two benefits for subject consistency: it reduces background interference and  computational cost by focusing on the subject alone. 
We adopt a similar strategy to that of previous methods~\cite{hertz2022prompt,tewel2024training}, using image-text cross-attention to extract subject masks. 
Specifically, let $X_\mathsf{id}$ denote the features of the reference image $\mathbf{x}_\mathsf{id}$ generated from the identity prompt $\mathbf{t}_\mathsf{id}$. When generating $\mathbf{x}_\mathsf{id}$, we average the cross-attention maps 
at the final denoising timestep for subject-related tokens (\eg, \texttt{``fairy''}), followed by applying Otsu’s method~\cite{otsu1975threshold} to produce a binary mask $M_\mathsf{id}$.
This mask highlights the subject-relevant regions, from which we extract the subject features as:
\begin{equation}
S_\mathsf{id} = X_\mathsf{id} \otimes M_\mathsf{id} \in \mathbb{R}^{s_\mathsf{id} \times d}
\end{equation}
where $\otimes$ applies the binary mask to retain subject features, $s_\mathsf{id}$ denotes the number of ones in the binary mask $M_\mathsf{id}$, and $d$ is the feature dimension.
Similarly, for each target image $\mathbf{x}_n$, we extract subject features as $S_n = X_n \otimes M_n \in \mathbb{R}^{s_n \times d} $. The process is visualized in Figure~\ref{fig:method}~(a).

\textbf{Transport between ${S}_\mathsf{id}$ and ${S}_{n}$.} Given the subject feature pairs $S_\mathsf{id}=[\mathbf{s}_\mathsf{id}^{1},...,\mathbf{s}_\mathsf{id}^{s_\mathsf{id}}]^\top$ and $S_n =[\mathbf{s}_n^{1},...,\mathbf{s}_n^{s_n}]^\top$, we first derive an optimal transport plan $T$ that aligns the reference features set $\{\mathbf{s}_\mathsf{id}^i\}_{i=1}^{s_\mathsf{id}}$ with the target features  $\{\mathbf{s}_n^i\}_{i=1}^{s_n}$ (See Figure~\ref{fig:method}~(b)). Using this plan $T$, we compose the target subject features by transporting features from the reference image. Intuitively, this process resembles \textbf{mosaicking}: we assemble the target subject using pieces from the reference image, \textbf{rearranged to match the target pose}. Since the visual pieces originate from the reference image, \textbf{subject identity is naturally preserved}.
To solve the OT problem in Eq.~\eqref{eq:ot}, we first define the cost matrix $C$ and the associated probability masses $\mathbf{a}$ and $\mathbf{b}$.

\noindent\textit{Definition of the cost matrix $C$}. The cost matrix is typically defined based on the pairwise distances between features: smaller distances imply lower transport costs. For a pair $\mathbf{s}_\mathsf{id}^i$ and $\mathbf{s}_n^j$  from final denoising step (where features contain minimal noise), the cost is defined as: 
\begin{equation}
    C(i,j)=1-\mathrm{cos}(\mathbf{s}_\mathsf{id}^i, \mathbf{s}_n^j)=1-\frac{{\mathbf{s}_\mathsf{id}^i}^\top\mathbf{s}_n^j }{\|\mathbf{s}_\mathsf{id}^i\|_2 \|\mathbf{s}_n^j\|_2}.
\end{equation}

\noindent\textit{Definition of the probability masses $\mathbf{a}$ and $\mathbf{b}$.} Intuitively, $\mathbf{a} = [a_1, \dots, a_{s_\mathsf{id}}]^\top$ represents the importance weights of the subject features, where a larger $a_i$ indicates that feature $\mathbf{s}_\mathsf{id}^i$ is more relevant to the subject $\mathbf{t}_\mathsf{id}$. We reuse the average cross-attention maps for generating the subject-relevant mask as the feature importance and apply softmax function to ensure the sum $\sum_i a_i$ equals to 1. The importance weights $\mathbf{b}$ for the target subject features $S_n$ are derived analogously.

With the cost matrix $C$ and the probability masses $\mathbf{a}$ and $\mathbf{b}$, we solve the OT plan $T_n$ in Eq.~\eqref{eq:ot} using network simplex algorithm~\cite{orlin1997polynomial}. With the derived $T_n$, the subject target features composed by reference subject features are computed as 
\begin{equation}
    S_{n}^\mathsf{OT}=T_n^\top S_\mathsf{id}.
\end{equation}
To form the final representation $X^\mathsf{OT}_n$, we combine $S_{n}^\mathsf{OT}$ with the non-subject features (masked out by $M_n$) from $X_n$. The representation is then passed through the diffusion network to produce the output. 
The \oursIT~process is illustrated in Figure~\ref{fig:method}~(c).

\subsection{Identity Refinement}
\label{section:stage2} 
The motivation behind this stage is that the \oursIT~module performs a coarse transport between $S_\mathsf{id}$ and $S_n$. However, since the binary subject masks are imprecise and the foreground of target images evolves during denoising—while our transport plan $T_n$ remains fixed—further refinement of subject details becomes necessary.

Our \oursIR~operates in the later denoising steps (\eg, the last 40 of 50 total steps) to reinforce subject details in the target images.
\oursIR~resembles cross-image attention-based SCG methods, except that each target image attend only to the most relevant reference features to avoid entangled feature update across target images. 
Specifically, to generate the $n$-th image, we first construct the concatenated keys and values as $K_{n,\mathsf{id}}=[K_n \oplus K_\mathsf{id}]$ and $V_{n,\mathsf{id}}=[V_n \oplus V_\mathsf{id}]$, respectively. The cross-image attention scores are compute as 
\begin{equation}
    A_n = \mathrm{softmax}\left(\frac{Q_nK_{n,\mathsf{id}}^\top}{\sqrt{d}}\right).
\end{equation}
For each query, we retain only the top-$\alpha$ attention scores of the reference tokens (\ie, $K_\mathsf{id}$). Specifically, for each row $A_i$ of $A$, we define the top-$\alpha$ index set $\mathcal{I}_i$ (see Appendix~\ref{A-implementation_details} for details) and zero out all other entries of $K_\mathsf{id}$:
\begin{equation}\label{eq:cross_attn_w}
    \tilde{A}_{ij} =
\begin{cases}
A_{ij}, & \text{if } j \in \mathcal{I}_i \\
0, & \text{otherwise}
\end{cases}
\quad \text{and} \quad
\hat{A}_i = \frac{\tilde{A}_i}{\sum_{j \in \mathcal{I}_i} \tilde{A}_{ij}}.
\end{equation}
The final cross-attention output is then computed as:
\begin{equation}\label{eq:alpha}
    \mathrm{Attn}_\alpha(Q_n,K_{n,\mathsf{id}}, V_{n,\mathsf{id}})=\hat{A}V_{n,\mathsf{id}}
\end{equation}
This filtering mechanism ensures that only the most relevant identity features from the reference image contribute to the attention update. The \oursIR~process is demonstrated in Figure~\ref{fig:method}~(d).
\begin{figure}[t]
  \centering
  \includegraphics[width=1.0\linewidth]{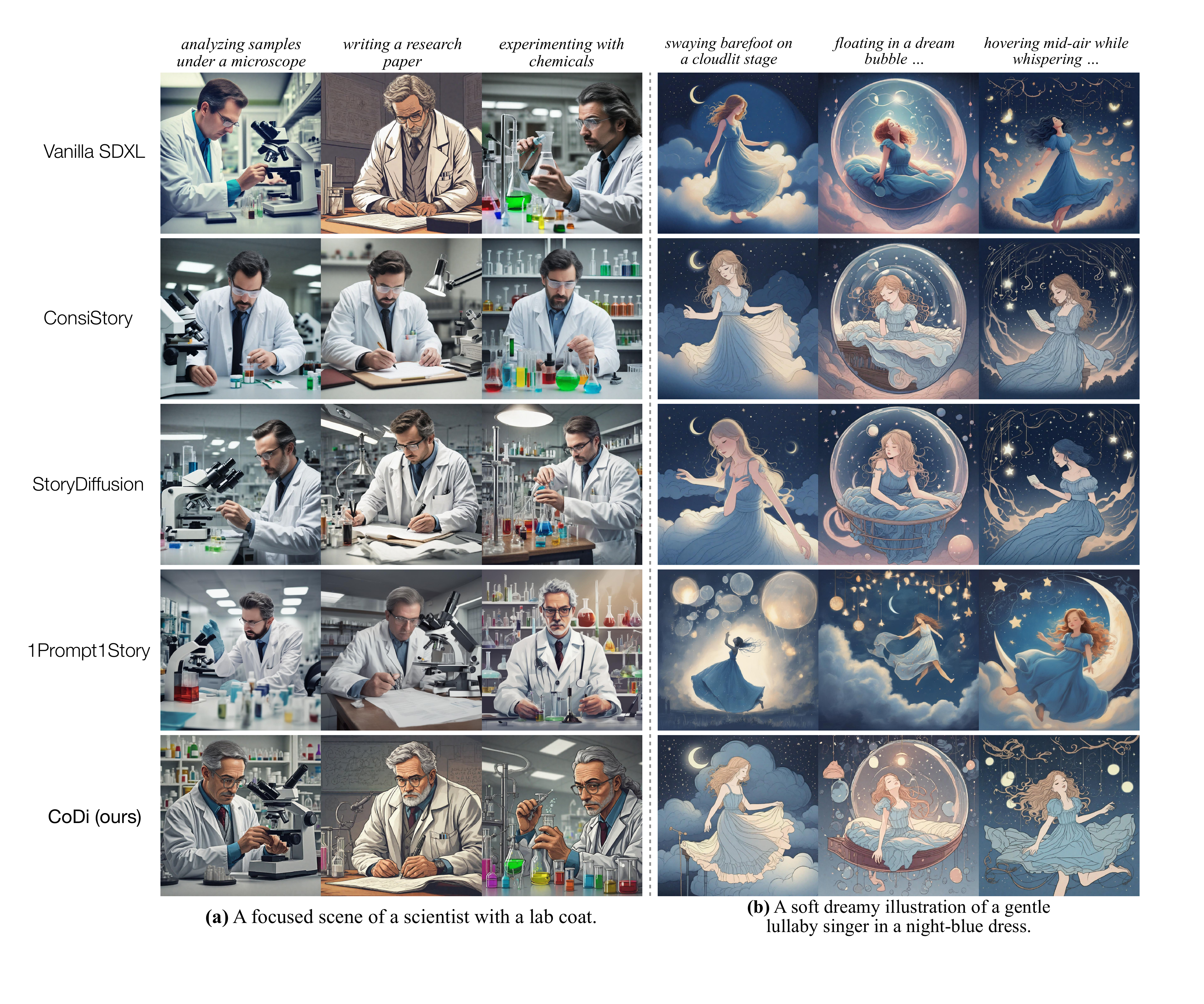}
  \caption{
  \textbf{Qualitative comparison} among Vanilla SDXL~\cite{podell2023sdxl}, ConsiStory~\cite{tewel2024training}, StoryDiffusion~\cite{zhou2024storydiffusion}, and 1Prompt1Story~\cite{liu2025one}.  ConsiStory and StoryDiffusion generate similar poses across examples, while 1Prompt1Story preserves pose diversity but struggles with subject consistency. In contrast, our \ours~achieves both.}
  \label{fig:combine_show}
\end{figure}
\begin{table}[t]
    \centering
    \caption{\textbf{Quantitative comparison} of subject consistency, pose diversity and prompt fidelity. Best results are marked in \textbf{bold}. $\uparrow$ indicates higher is better, and $\downarrow$ indicates lower is better.}
    \label{tab:results}
    \tabstyle{7pt}
    \begin{adjustbox}{max width=\linewidth}
    \begin{tabular}{l|ccc|c|c}
    \toprule
    \multirow{2}{*}{{Method}} & \multicolumn{3}{c|}{ \textbf{Subject Consistency}} & \textbf{Pose }   & \textbf{Prompt }  \\ % \cmidrule(lr){2-4} 
                   & CLIP-I ($\uparrow$) & DINO-v2 ($\uparrow$) & DreamSim ($\downarrow$) & \textbf{Diversity} ($\uparrow$) &  \textbf{Fidelity} ($\uparrow$) \\
    \midrule
    % Vanilla SD1.5 & 0.7929 &0.7558 &0.4371 & & 0.8525 \\
    Vanilla SDXL~\cite{podell2023sdxl} & 0.8417 & 0.8010 & 0.3139 & 0.0772 & 0.9082 \\
    1Prompt1Story~\cite{liu2025one} & 0.8627 & 0.8233 & 0.2959 & 0.0662 & 0.8814 \\
    ConsiStory~\cite{tewel2024training} & 0.8751 & 0.8428 & 0.2336 & 0.0621 & \textbf{0.9148} \\
    StoryDiffusion~\cite{zhou2024storydiffusion} & 0.8776 & 0.8471 & 0.2356 & 0.0605 & 0.9038 \\
 \midrule
        \rowcolor{lightCyan}
    \ours~{(ours)} & \textbf{0.8809} & \textbf{0.8514} & \textbf{0.2136} & \textbf{0.0758} & 0.9041 \\
    \bottomrule
    \end{tabular}
    \end{adjustbox}
\end{table}

\section{Experiments}
\subsection{Setup}\label{sec:expsetup}
\noindent\textbf{Benchmark.} 
We evaluate our \ours~on the standard SCG benchmark, ConsiStory+\cite{liu2025one}, which comprises nearly 200 prompt sets and supports the generation of over 1,100 images. Each prompt set includes a subject described in a specific style, with multiple frame-specific descriptions.

\noindent\textbf{Baselines and implementation details.} We compare our \ours~with SoTA training-free SCG methods, including ConsiStory~\cite{tewel2024training}, StoryDiffusion~\cite{zhou2024storydiffusion} and 1Prompt1Story~\cite{liu2025one}. 
We reproduce all baselines using their official released code.
All methods are implemented using the same backbone model, Stable Diffusion XL 1.0~\cite{podell2023sdxl}, with an image resolution of $1024 \times 1024$, {except for StoryDiffusion, which is evaluated at $768 \times 768$ due to its high memory consumption, following its original setting}.
% To ensure fairness, identical noise seeds are used for all methods. 
To ensure fairness, identical noise seeds are enforced for all methods, ensuring that each prompt is initialized with the same random noise input.
% We set the hyperparameter $\alpha$ in Eq.~\eqref{eq:alpha} to select the top 50\% of reference features.
Hyperparameter $\alpha$ in Eq.~\eqref{eq:alpha} selects the top 50\% of reference features.

\noindent\textbf{Evaluation metrics.}
Our evaluation framework assesses the quality of generated images from three aspects: \textbf{(1)} subject consistency, \textbf{(2)} pose diversity, and \textbf{(3)} prompt fidelity. Subject consistency is evaluated by computing the average pairwise cosine similarity (or distance) between image embeddings within each target image set. We use three image encoders for this evaluation: CLIP-I~\cite{hessel2021clipscore}, DINO-v2~\cite{oquab2023dinov2}, and DreamSim~\cite{fu2023dreamsim}. 
To evaluate pose diversity, we extract 2D human joint coordinates using ViTPose's pose estimation model~\cite{xu2022vitpose}. To eliminate global variations in translation, rotation, and scale, we align poses using Procrustes analysis~\cite{schonemann1966generalized}, inspired by standard practices in face alignment~\cite{9442331}.
The pose diversity score is then computed as the average Euclidean distance between corresponding keypoints across aligned image pairs. A higher score indicates greater pose diversity.
% For prompt fidelity, we use CLIP-Score~\cite{hessel2021clipscore} to measure the cosine similarity of embeddings between each image and its corresponding textual prompt.
For prompt fidelity, we use CLIP-Score~\cite{hessel2021clipscore} to measure the cosine similarity between image and textual prompt embeddings.
See Appendix~\ref{A-evaluation_details} for more details.
% See Appendix A.1 for more details.

\subsection{Experimental Results}
\noindent\textbf{Qualitative comparison.} 
As shown in Figure~\ref{fig:combine_show}, our \ours~achieves superior visual quality in terms of pose diversity, subject consistency, and prompt fidelity.
Our \ours~preserves the pose diversity of Vanilla SDXL~\cite{podell2023sdxl} while overcoming its limitation in subject consistency. 
In comparison, ConsiStory~\cite{tewel2024training} and StoryDiffusion~\cite{zhou2024storydiffusion} achieve subject consistency at the cost of pose diversity. For example, in the scientist scenario, the man exhibits nearly identical body poses.
Although 1Prompt1Story~\cite{liu2025one} maintains strong layout and pose diversity in both cases, its subject consistency remains limited.

\begin{figure}[t]
  \centering
  \includegraphics[width=1.0\linewidth]{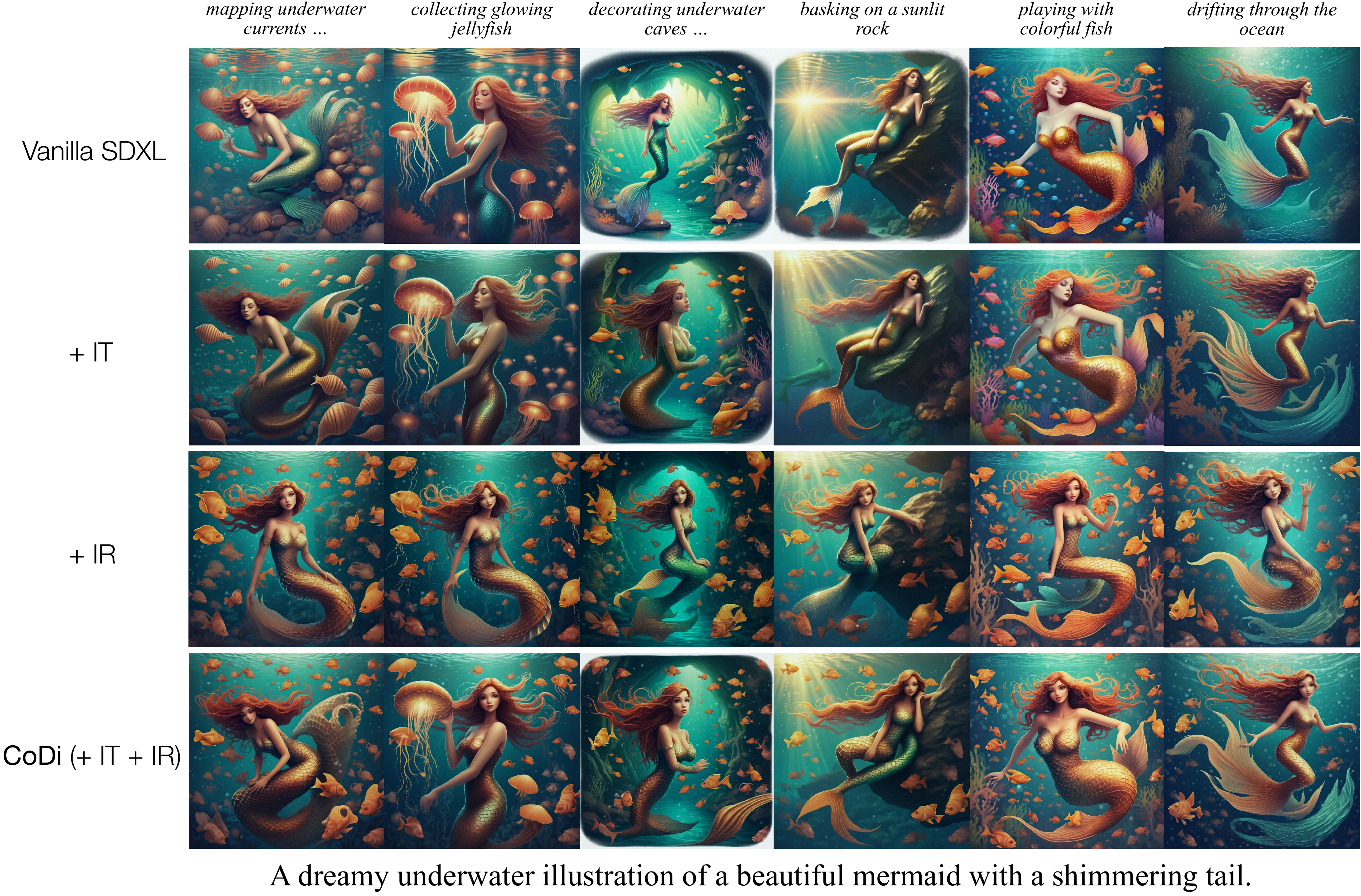}
  \caption{
  \textbf{Main component analysis (qualitative)} on identity transport (\oursIT) and identity refinement (\oursIR).
   \oursIT~enhances subject consistency in the \textit{coarse-grained level} and preserves pose diversity. \oursIR~enhances subject consistency in the \textit{fine-grained level} reduces pose diversity. Their combination yields the best consistency and preserves diversity.
  }
  \label{fig:ablation}
\end{figure}
\begin{table}[t]
    \centering
    % \vspace{-0.5cm}
    \caption{
    \textbf{Main component analysis (quantitative)} on identity transport (\oursIT) and identity refinement (\oursIR). \oursIT~enhances subject consistency  and preserves pose diversity. \oursIR~enhances subject consistency while reduces pose diversity. Their combination yields the best consistency and preserves diversity.}
    \label{tab:ablation}
        \tabstyle{10pt}
    \begin{tabular}{cc|ccc|c|c}
    \toprule
    \multirow{2}{*}{\oursIT} & \multirow{2}{*}{\oursIR} & \multicolumn{3}{c|}{ \textbf{Subject Consistency}} & \textbf{Pose} & \textbf{Prompt} \\
     &  & {CLIP-I} ($\uparrow$) & {DINO-v2} ($\uparrow$) & {DreamSim} ($\downarrow$) & \textbf{Diversity} ($\uparrow$) & \textbf{Fidelity} ($\uparrow$) \\
    \midrule
     &  & 0.8417 & 0.8010 & 0.3139 & 0.0772 & 0.9082 \\
     \checkmark &  & 0.8576 & 0.8207 & 0.2707 & 0.0800 & 0.9090 \\
     &\checkmark  & 0.8859 & 0.8618 & 0.2044 & 0.0675 & 0.8975 \\
    \checkmark & \checkmark & 0.8809 & 0.8514 & 0.2136 & 0.0758 & 0.9041 \\
    \bottomrule
    \end{tabular}
    % \vspace{-0.2cm}
\end{table} 

\noindent\textbf{Quantitative comparison.}
Table~\ref{tab:results} presents a quantitative comparison. 
\textbf{(1)} Across all three subject consistency metrics—CLIP-I~\cite{hessel2021clipscore}, DINO-v2~\cite{oquab2023dinov2}, and DreamSim~\cite{fu2023dreamsim}—our \ours~achieves the best performance, demonstrating superior identity preservation across instances.
In particular, our method obtains the lowest DreamSim score (0.2136), indicating closer alignment with human perceptual similarity than competing methods.
\textbf{(2)} 
In terms of pose diversity, \ours~achieves the highest score (0.0758), closely matching Vanilla SDXL (0.0772). This demonstrates its ability to preserve the inherent pose diversity of the diffusion model while maintaining subject consistency.
\textbf{(3)}
For prompt fidelity, \ours~performs competitively—ranking second only to ConsiStory and comparable to Vanilla SDXL.
These results demonstrate \ours's ability to achieve subject consistency without compromising pose diversity or prompt alignment.

\begin{figure}[t]
  \centering
  \includegraphics[width=1.0\linewidth]{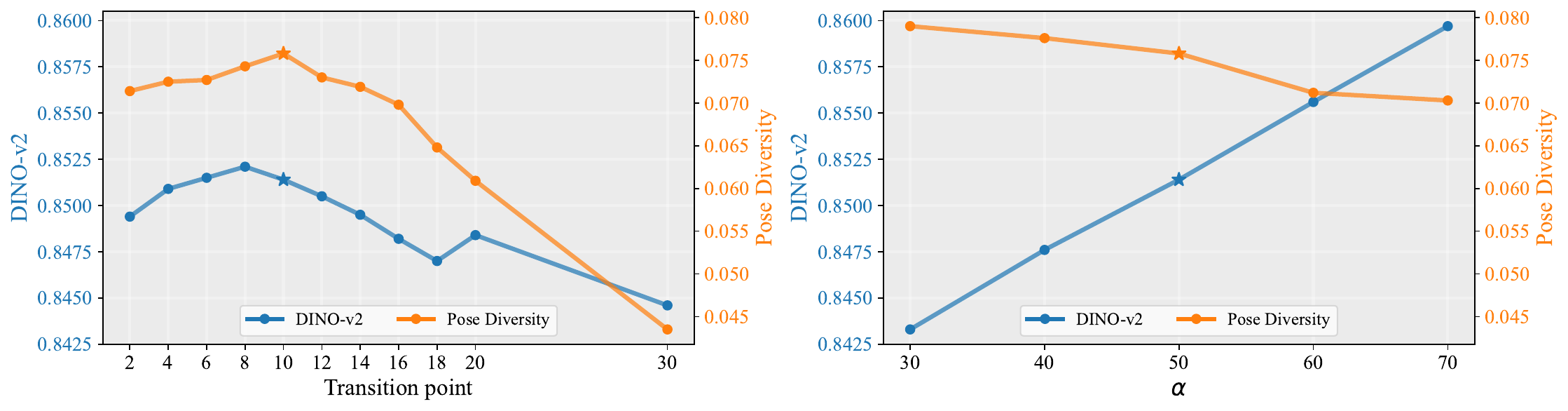}
  \caption{
  Ablation studies on \textbf{(a)} stage transition point, and \textbf{(b)} the effect of $\alpha$.
  }
  \label{fig:combine_ablate}
  % \vspace{-0.6cm}
\end{figure}

\subsection{Ablation Studies}

\noindent\textbf{Main component analysis.}
The contribution of each module (\oursIT~and \oursIR) to subject consistency and pose diversity are evaluated through quantitative and qualitative ablations, as shown in Table~\ref{tab:ablation} and Figure~\ref{fig:ablation}.
Table~\ref{tab:ablation} shows that both \oursIT~and \oursIR~improve subject consistency, while \oursIT~also enhances pose diversity. However, using \oursIR~alone reduces pose diversity—for example, the score drops from 0.0772 to 0.0675 compared to the SDXL baseline.
When both modules are applied, subject consistency further improves due to their synergistic effect, while pose diversity is preserved.

Figure~\ref{fig:ablation} visualizes the effect of each module. Compared to Vanilla SDXL, applying \oursIT~preserves the original pose and improves subject consistency, but some details, such as facial identity, remain suboptimal. 
% In contrast, \oursIR~alone enhances fine-grained consistency, but produces nearly identical poses across images, resulting in a substantial loss of diversity.
In contrast, \oursIR~alone enhances fine-grained consistency, but produces nearly identical poses across images, substantially reducing diversity.
As shown in the bottom row, combining \oursIT~and \oursIR~improves both coarse and fine-grained consistency without compromising pose diversity.

\noindent\textbf{Study on stage transition point.}
Our \ours~adopts a two-stage strategy: identity transport (\oursIT) in the early denoising steps and identity refinement (\oursIR) in the later ones. By default, we set the stage transition point at step $t=10$ out of a total of 50 denoising steps (\oursIT~is applied when $t \leq 10$, and \oursIR~afterward).
In this study, we investigate how the choice of transition point affects generation quality. As shown in Figure~\ref{fig:combine_ablate}~(a), we vary $t$ from 2 to 30 and evaluate subject consistency (DINO-v2) and pose diversity.
We find that our default choice $t=10$ achieves a favorable trade-off between consistency and diversity.

\noindent\textbf{Effect of $\alpha$.} 
In the \oursIR~stage, we select the top-$\alpha$ percent of reference features to inject into the target subject features.
In this study, we examine how varying $\alpha$ affects subject-consistent generation. Specifically, we vary $\alpha$ from 30\% to 70\% and report subject consistency (DINO-v2) and pose diversity in Figure~\ref{fig:combine_ablate}~(b).
We observe that increasing $\alpha$ improves subject consistency but reduces pose diversity. Setting $\alpha=50\%$ provides a favorable trade-off.

\subsection{Multi-subject Generation}

\begin{figure}[t]
  % \vspace{-1.3cm}
  \centering
  \includegraphics[width=1.0\linewidth]{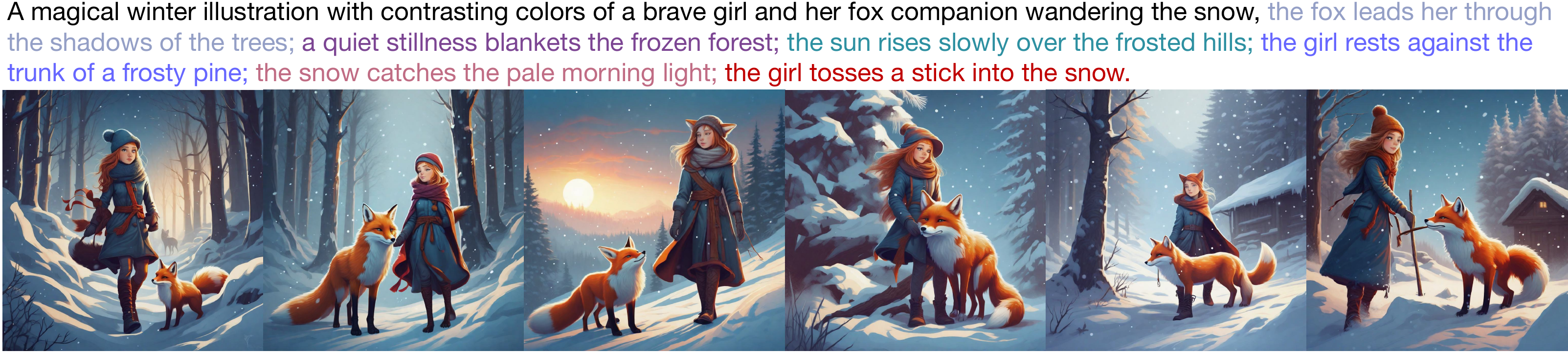}
  \caption{Multi-subject generation. \ours~consistently preserves identities of multiple subjects while maintaining diverse poses and spatial layouts.}
  \label{A-fig:mul_sub}
  % \vspace{-0.3cm}
\end{figure}

Both~\oursIT~and~\oursIR~are independently applied to each subject and are easily extendable, enabling our~\ours~to naturally support multi-subject generation. For each subject, we perform \oursIT~and extract its most salient features for \oursIR~, which effectively prevents feature interference across subjects and enhances subject consistency. As shown in Fig.~\ref{A-fig:mul_sub}, our~\ours~preserves multi-subject consistency while maintaining their pose diversity.

\subsection{Different Style Generation}
\begin{figure}[t]
  \centering
  \includegraphics[width=1.0\linewidth]{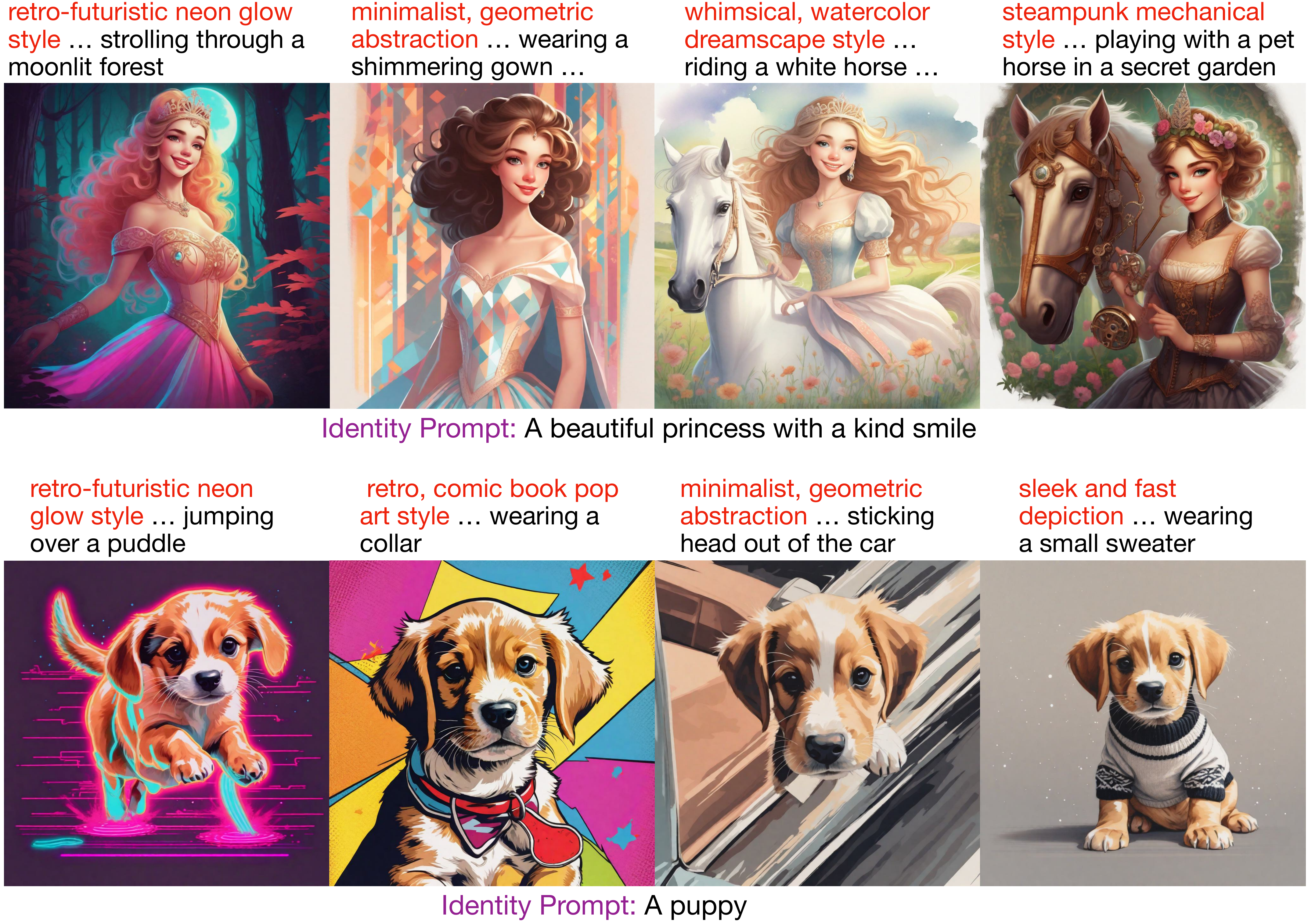}
  \caption{\ours~can generate images with consistent subject appearance and different style.}
  \label{A-fig:different_style}
\end{figure}

\ours~first transports the identity features from the reference image during the IT stage, and in the IR stage, the diffusion model refines the subject with a specific style. As shown in Fig.~\ref{A-fig:different_style}, CoDi generates images with consistent subject appearance and diverse styles.

\subsection{User Study}

We conducted a user study to compare our method with state-of-the-art approaches. 
A total of 30 prompt sets were randomly sampled, each consisting of four fixed-length prompts. Thirty-nine participants were asked to evaluate which method demonstrated the best overall performance of the generated images in terms of subject consistency, pose/layout diversity, and prompt fidelity. 
As shown in Table~\ref{tab:user_study}, \ours~achieved the highest overall preference, surpassing the second-best method (StoryDiffusion) by 24.01\%.

\begin{table}[t]
\centering
\caption{User study with 39 participants evaluating T2I SCG methods based on human preference.}
\begin{tabular}{l c c c >{\columncolor{lightCyan}}c}
\toprule
Method & ConsiStory & StoryDiffusion & 1Prompt1Story & \ours~(ours) \\
\midrule
Percent (\%) $\uparrow$ & 19.06 & 21.20 & 14.53 & 45.21 \\
\bottomrule
\end{tabular}

\label{tab:user_study}
\end{table}

\section{Conclusion}

In this paper, we propose \ours, a novel training-free framework that addresses the trade-off between subject consistency and pose diversity.
\ours~comprises two key components: identity transport (\oursIT) and identity refinement (\oursIR). 
During early denoising steps, \oursIT~aligns features across instances by optimally transporting the identity subject’s features, % to each target image, 
while preserving pose diversity.
\oursIR~further refines subject consistency by aligning instance features with the salient attributes of the identity subject in the later denoising steps.
The effectiveness of our \ours~is demonstrated by its state-of-the-art performance in achieving subject consistency and maintaining pose diversity.

% subsubsection
\section*{Acknowledgments} 
This work was supported by the Gusu Innovation and Entrepreneur Leading Talents: No. ZXL2024362, Natural Science Foundation of Jiangsu Province: BK20241198, and Natural Science Foundation of China: No. 62406135.

\bibliography{iclr2026_conference}

\begin{thebibliography}{55}
\providecommand{\natexlab}[1]{#1}
\providecommand{\url}[1]{\texttt{#1}}
\expandafter\ifx\csname urlstyle\endcsname\relax
  \providecommand{\doi}[1]{doi: #1}\else
  \providecommand{\doi}{doi: \begingroup \urlstyle{rm}\Url}\fi

\bibitem[Avrahami et~al.(2024)Avrahami, Hertz, Vinker, Arar, Fruchter, Fried, Cohen-Or, and Lischinski]{avrahami2024chosen}
Omri Avrahami, Amir Hertz, Yael Vinker, Moab Arar, Shlomi Fruchter, Ohad Fried, Daniel Cohen-Or, and Dani Lischinski.
\newblock The chosen one: Consistent characters in text-to-image diffusion models.
\newblock In \emph{ACM SIGGRAPH}, 2024.

\bibitem[Betker et~al.(2023)Betker, Goh, Jing, Brooks, Wang, Li, Ouyang, Zhuang, Lee, Guo, et~al.]{betker2023improving}
James Betker, Gabriel Goh, Li~Jing, Tim Brooks, Jianfeng Wang, Linjie Li, Long Ouyang, Juntang Zhuang, Joyce Lee, Yufei Guo, et~al.
\newblock Improving image generation with better captions.
\newblock \emph{Computer Science. https://cdn. openai. com/papers/dall-e-3. pdf}, 2\penalty0 (3):\penalty0 8, 2023.

\bibitem[Blattmann et~al.(2023)Blattmann, Rombach, Ling, Dockhorn, Kim, Fidler, and Kreis]{blattmann2023align}
Andreas Blattmann, Robin Rombach, Huan Ling, Tim Dockhorn, Seung~Wook Kim, Sanja Fidler, and Karsten Kreis.
\newblock Align your latents: High-resolution video synthesis with latent diffusion models.
\newblock In \emph{CVPR}, 2023.

\bibitem[Chen et~al.(2025)Chen, Wang, Pan, Zhu, Yang, and Zhang]{chen2025detailtrainingfreeenhancertexttoimage}
Lifeng Chen, Jiner Wang, Zihao Pan, Beier Zhu, Xiaofeng Yang, and Chi Zhang.
\newblock Detail++: Training-free detail enhancer for text-to-image diffusion models, 2025.
\newblock URL \url{https://arxiv.org/abs/2507.17853}.

\bibitem[Esser et~al.(2024)Esser, Kulal, Blattmann, Entezari, M{\"u}ller, Saini, Levi, Lorenz, Sauer, Boesel, et~al.]{esser2024scaling}
Patrick Esser, Sumith Kulal, Andreas Blattmann, Rahim Entezari, Jonas M{\"u}ller, Harry Saini, Yam Levi, Dominik Lorenz, Axel Sauer, Frederic Boesel, et~al.
\newblock Scaling rectified flow transformers for high-resolution image synthesis.
\newblock In \emph{ICML}, 2024.

\bibitem[Fu et~al.(2023)Fu, Tamir, Sundaram, Chai, Zhang, Dekel, and Isola]{fu2023dreamsim}
Stephanie Fu, Netanel Tamir, Shobhita Sundaram, Lucy Chai, Richard Zhang, Tali Dekel, and Phillip Isola.
\newblock Dreamsim: Learning new dimensions of human visual similarity using synthetic data.
\newblock In \emph{NeurIPS}, 2023.

\bibitem[Gal et~al.(2023{\natexlab{a}})Gal, Alaluf, Atzmon, Patashnik, Bermano, Chechik, and Cohen-Or]{gal2022image}
Rinon Gal, Yuval Alaluf, Yuval Atzmon, Or~Patashnik, Amit~H Bermano, Gal Chechik, and Daniel Cohen-Or.
\newblock An image is worth one word: Personalizing text-to-image generation using textual inversion.
\newblock In \emph{ICLR}, 2023{\natexlab{a}}.

\bibitem[Gal et~al.(2023{\natexlab{b}})Gal, Arar, Atzmon, Bermano, Chechik, and Cohen-Or]{gal2023designing}
Rinon Gal, Moab Arar, Yuval Atzmon, Amit~H Bermano, Gal Chechik, and Daniel Cohen-Or.
\newblock Designing an encoder for fast personalization of text-to-image models.
\newblock \emph{arXiv preprint arXiv:2302.12228}, 2\penalty0 (3), 2023{\natexlab{b}}.

\bibitem[Hertz et~al.(2023)Hertz, Mokady, Tenenbaum, Aberman, Pritch, and Cohen-Or]{hertz2022prompt}
Amir Hertz, Ron Mokady, Jay Tenenbaum, Kfir Aberman, Yael Pritch, and Daniel Cohen-Or.
\newblock Prompt-to-prompt image editing with cross attention control.
\newblock In \emph{ICLR}, 2023.

\bibitem[Hertz et~al.(2024)Hertz, Voynov, Fruchter, and Cohen-Or]{Hertz_2024_CVPR}
Amir Hertz, Andrey Voynov, Shlomi Fruchter, and Daniel Cohen-Or.
\newblock Style aligned image generation via shared attention.
\newblock In \emph{CVPR}, 2024.

\bibitem[Hessel et~al.(2021)Hessel, Holtzman, Forbes, Bras, and Choi]{hessel2021clipscore}
Jack Hessel, Ari Holtzman, Maxwell Forbes, Ronan~Le Bras, and Yejin Choi.
\newblock Clipscore: A reference-free evaluation metric for image captioning.
\newblock In \emph{EMNLP}, 2021.

\bibitem[Hu et~al.(2023)Hu, Zhong, Liang, Zhang, Li, Li, and Ji]{hu2023transformer}
Xiantao Hu, Bineng Zhong, Qihua Liang, Shengping Zhang, Ning Li, Xianxian Li, and Rongrong Ji.
\newblock Transformer tracking via frequency fusion.
\newblock \emph{TCSVT}, 2023.

\bibitem[Kopiczko et~al.(2024)Kopiczko, Blankevoort, and Asano]{kopiczko2023vera}
Dawid~J Kopiczko, Tijmen Blankevoort, and Yuki~M Asano.
\newblock Vera: Vector-based random matrix adaptation.
\newblock In \emph{ICLR}, 2024.

\bibitem[Kumari et~al.(2023)Kumari, Zhang, Zhang, Shechtman, and Zhu]{kumari2023multi}
Nupur Kumari, Bingliang Zhang, Richard Zhang, Eli Shechtman, and Jun-Yan Zhu.
\newblock Multi-concept customization of text-to-image diffusion.
\newblock In \emph{CVPR}, 2023.

\bibitem[Lei et~al.(2025)Lei, Song, Zhu, Wang, and Zhang]{Lei_2025_CVPR}
Mingkun Lei, Xue Song, Beier Zhu, Hao Wang, and Chi Zhang.
\newblock Stylestudio: Text-driven style transfer with selective control of style elements.
\newblock In \emph{CVPR}, 2025.

\bibitem[Li et~al.(2019)Li, Gan, Shen, Liu, Cheng, Wu, Carin, Carlson, and Gao]{li2019storygan}
Yitong Li, Zhe Gan, Yelong Shen, Jingjing Liu, Yu~Cheng, Yuexin Wu, Lawrence Carin, David Carlson, and Jianfeng Gao.
\newblock Storygan: A sequential conditional gan for story visualization.
\newblock In \emph{CVPR}, 2019.

\bibitem[Li et~al.(2024)Li, Cao, Wang, Qi, Cheng, and Shan]{li2024photomaker}
Zhen Li, Mingdeng Cao, Xintao Wang, Zhongang Qi, Ming-Ming Cheng, and Ying Shan.
\newblock Photomaker: Customizing realistic human photos via stacked id embedding.
\newblock In \emph{CVPR}, 2024.

\bibitem[Lin et~al.(2021)Lin, Zhu, Wang, Liao, Qian, Lu, and Zhou]{9442331}
Chunze Lin, Beier Zhu, Quan Wang, Renjie Liao, Chen Qian, Jiwen Lu, and Jie Zhou.
\newblock Structure-coherent deep feature learning for robust face alignment.
\newblock \emph{IEEE Transactions on Image Processing}, 30:\penalty0 5313--5326, 2021.

\bibitem[Lin et~al.(2025{\natexlab{a}})Lin, Wang, Ren, and Han]{lin2025pancap}
Kun-Yu Lin, Hongjun Wang, Weining Ren, and Kai Han.
\newblock Panoptic captioning: An equivalence bridge for image and text.
\newblock In \emph{The Thirty-Ninth Annual Conference on Neural Information Processing Systems}, 2025{\natexlab{a}}.

\bibitem[Lin et~al.(2025{\natexlab{b}})Lin, Liang, Hu, Lin, Kang, Tian, Lai, and Zheng]{lin2025coopdiff}
Xiaotong Lin, Tianming Liang, Jian-Fang Hu, Kun-Yu Lin, Yulei Kang, Chunwei Tian, Jianhuang Lai, and Wei-Shi Zheng.
\newblock Coopdiff: Anticipating 3d human-object interactions via contact-consistent decoupled diffusion.
\newblock \emph{arXiv preprint arXiv:2508.07162}, 2025{\natexlab{b}}.

\bibitem[Liu et~al.(2024)Liu, Wu, Zhong, Zhang, Wang, and Xie]{liu2024intelligent}
Chang Liu, Haoning Wu, Yujie Zhong, Xiaoyun Zhang, Yanfeng Wang, and Weidi Xie.
\newblock Intelligent grimm-open-ended visual storytelling via latent diffusion models.
\newblock In \emph{CVPR}, 2024.

\bibitem[Liu et~al.(2025)Liu, Wang, Li, van~de Weijer, Khan, Yang, Wang, Yang, and Cheng]{liu2025one}
Tao Liu, Kai Wang, Senmao Li, Joost van~de Weijer, Fahad~Shahbaz Khan, Shiqi Yang, Yaxing Wang, Jian Yang, and Ming-Ming Cheng.
\newblock One-prompt-one-story: Free-lunch consistent text-to-image generation using a single prompt.
\newblock In \emph{ICLR}, 2025.

\bibitem[Maharana et~al.(2022)Maharana, Hannan, and Bansal]{maharana2022storydall}
Adyasha Maharana, Darryl Hannan, and Mohit Bansal.
\newblock Storydall-e: Adapting pretrained text-to-image transformers for story continuation.
\newblock In \emph{ECCV}. Springer, 2022.

\bibitem[Mei et~al.(2025)Mei, Talebi, Ardakani, Patel, Milanfar, and Delbracio]{mei2025power}
Kangfu Mei, Hossein Talebi, Mojtaba Ardakani, Vishal~M Patel, Peyman Milanfar, and Mauricio Delbracio.
\newblock The power of context: How multimodality improves image super-resolution.
\newblock In \emph{CVPR}, 2025.

\bibitem[Monge(1781)]{monge1781memoire}
Gaspard Monge.
\newblock M{\'e}moire sur la th{\'e}orie des d{\'e}blais et des remblais.
\newblock \emph{Mem. Math. Phys. Acad. Royale Sci.}, pp.\  666--704, 1781.

\bibitem[Mou et~al.(2024)Mou, Wang, Xie, Wu, Zhang, Qi, and Shan]{mou2024t2i}
Chong Mou, Xintao Wang, Liangbin Xie, Yanze Wu, Jian Zhang, Zhongang Qi, and Ying Shan.
\newblock T2i-adapter: Learning adapters to dig out more controllable ability for text-to-image diffusion models.
\newblock In \emph{AAAI}, 2024.

\bibitem[Nan et~al.(2025)Nan, Xie, Zhou, Fan, Yang, Chen, Li, Yang, and Tai]{nan2024openvid}
Kepan Nan, Rui Xie, Penghao Zhou, Tiehan Fan, Zhenheng Yang, Zhijie Chen, Xiang Li, Jian Yang, and Ying Tai.
\newblock Openvid-1m: A large-scale high-quality dataset for text-to-video generation.
\newblock In \emph{ICLR}, 2025.

\bibitem[Oquab et~al.(2023)Oquab, Darcet, Moutakanni, Vo, Szafraniec, Khalidov, Fernandez, Haziza, Massa, El-Nouby, et~al.]{oquab2023dinov2}
Maxime Oquab, Timoth{\'e}e Darcet, Th{\'e}o Moutakanni, Huy Vo, Marc Szafraniec, Vasil Khalidov, Pierre Fernandez, Daniel Haziza, Francisco Massa, Alaaeldin El-Nouby, et~al.
\newblock Dinov2: Learning robust visual features without supervision.
\newblock \emph{arXiv preprint arXiv:2304.07193}, 2023.

\bibitem[Orlin(1997)]{orlin1997polynomial}
James~B Orlin.
\newblock A polynomial time primal network simplex algorithm for minimum cost flows.
\newblock \emph{Mathematical Programming}, 78:\penalty0 109--129, 1997.

\bibitem[Otsu et~al.(1975)]{otsu1975threshold}
Nobuyuki Otsu et~al.
\newblock A threshold selection method from gray-level histograms.
\newblock \emph{Automatica}, 11\penalty0 (285-296):\penalty0 23--27, 1975.

\bibitem[Podell et~al.(2023)Podell, English, Lacey, Blattmann, Dockhorn, M{\"u}ller, Penna, and Rombach]{podell2023sdxl}
Dustin Podell, Zion English, Kyle Lacey, Andreas Blattmann, Tim Dockhorn, Jonas M{\"u}ller, Joe Penna, and Robin Rombach.
\newblock Sdxl: Improving latent diffusion models for high-resolution image synthesis.
\newblock \emph{arXiv preprint arXiv:2307.01952}, 2023.

\bibitem[Rahman et~al.(2023)Rahman, Lee, Ren, Tulyakov, Mahajan, and Sigal]{rahman2023make}
Tanzila Rahman, Hsin-Ying Lee, Jian Ren, Sergey Tulyakov, Shweta Mahajan, and Leonid Sigal.
\newblock Make-a-story: Visual memory conditioned consistent story generation.
\newblock In \emph{CVPR}, 2023.

\bibitem[Ramesh et~al.(2022)Ramesh, Dhariwal, Nichol, Chu, and Chen]{ramesh2022hierarchical}
Aditya Ramesh, Prafulla Dhariwal, Alex Nichol, Casey Chu, and Mark Chen.
\newblock Hierarchical text-conditional image generation with clip latents.
\newblock \emph{arXiv preprint arXiv:2204.06125}, 1\penalty0 (2):\penalty0 3, 2022.

\bibitem[Roich et~al.(2022)Roich, Mokady, Bermano, and Cohen-Or]{roich2022pivotal}
Daniel Roich, Ron Mokady, Amit~H Bermano, and Daniel Cohen-Or.
\newblock Pivotal tuning for latent-based editing of real images.
\newblock \emph{ACM Transactions on graphics (TOG)}, 42\penalty0 (1):\penalty0 1--13, 2022.

\bibitem[Rombach et~al.(2022)Rombach, Blattmann, Lorenz, Esser, and Ommer]{rombach2022high}
Robin Rombach, Andreas Blattmann, Dominik Lorenz, Patrick Esser, and Bj{\"o}rn Ommer.
\newblock High-resolution image synthesis with latent diffusion models.
\newblock In \emph{CVPR}, 2022.

\bibitem[Ruiz et~al.(2023)Ruiz, Li, Jampani, Pritch, Rubinstein, and Aberman]{ruiz2023dreambooth}
Nataniel Ruiz, Yuanzhen Li, Varun Jampani, Yael Pritch, Michael Rubinstein, and Kfir Aberman.
\newblock Dreambooth: Fine tuning text-to-image diffusion models for subject-driven generation.
\newblock In \emph{CVPR}, 2023.

\bibitem[Ruiz et~al.(2024)Ruiz, Li, Jampani, Wei, Hou, Pritch, Wadhwa, Rubinstein, and Aberman]{ruiz2024hyperdreambooth}
Nataniel Ruiz, Yuanzhen Li, Varun Jampani, Wei Wei, Tingbo Hou, Yael Pritch, Neal Wadhwa, Michael Rubinstein, and Kfir Aberman.
\newblock Hyperdreambooth: Hypernetworks for fast personalization of text-to-image models.
\newblock In \emph{CVPR}, 2024.

\bibitem[Saharia et~al.(2022)Saharia, Chan, Saxena, Li, Whang, Denton, Ghasemipour, Gontijo~Lopes, Karagol~Ayan, Salimans, et~al.]{saharia2022photorealistic}
Chitwan Saharia, William Chan, Saurabh Saxena, Lala Li, Jay Whang, Emily~L Denton, Kamyar Ghasemipour, Raphael Gontijo~Lopes, Burcu Karagol~Ayan, Tim Salimans, et~al.
\newblock Photorealistic text-to-image diffusion models with deep language understanding.
\newblock In \emph{NeurIPS}, 2022.

\bibitem[Sch{\"o}nemann(1966)]{schonemann1966generalized}
Peter~H Sch{\"o}nemann.
\newblock A generalized solution of the orthogonal procrustes problem.
\newblock \emph{Psychometrika}, 31\penalty0 (1):\penalty0 1--10, 1966.

\bibitem[Tewel et~al.(2024)Tewel, Kaduri, Gal, Kasten, Wolf, Chechik, and Atzmon]{tewel2024training}
Yoad Tewel, Omri Kaduri, Rinon Gal, Yoni Kasten, Lior Wolf, Gal Chechik, and Yuval Atzmon.
\newblock Training-free consistent text-to-image generation.
\newblock \emph{SIGGRAPH}, 2024.

\bibitem[Villani et~al.(2008)]{villani2008optimal}
C{\'e}dric Villani et~al.
\newblock \emph{Optimal transport: old and new}, volume 338.
\newblock Springer, 2008.

\bibitem[Wang et~al.(2025{\natexlab{a}})Wang, Li, Zhu, Yuan, Zhang, Yang, Chang, and Zhang]{wang2025paralleldiffusionsolverresidual}
Ruoyu Wang, Ziyu Li, Beier Zhu, Liangyu Yuan, Hanwang Zhang, Xun Yang, Xiaojun Chang, and Chi Zhang.
\newblock Parallel diffusion solver via residual dirichlet policy optimization, 2025{\natexlab{a}}.
\newblock URL \url{https://arxiv.org/abs/2512.22796}.

\bibitem[Wang et~al.(2025{\natexlab{b}})Wang, Zhu, Li, Yuan, and Zhang]{wang2025adaptive}
Ruoyu Wang, Beier Zhu, Junzhi Li, Liangyu Yuan, and Chi Zhang.
\newblock Adaptive stochastic coefficients for accelerating diffusion sampling.
\newblock In \emph{NeurIPS}, 2025{\natexlab{b}}.

\bibitem[Wu et~al.(2025)Wu, Huang, Wu, Cheng, Ding, and He]{wu2025less}
Shaojin Wu, Mengqi Huang, Wenxu Wu, Yufeng Cheng, Fei Ding, and Qian He.
\newblock Less-to-more generalization: Unlocking more controllability by in-context generation.
\newblock \emph{arXiv preprint arXiv:2504.02160}, 2025.

\bibitem[Xiao et~al.(2024)Xiao, Yin, Freeman, Durand, and Han]{xiao2024fastcomposer}
Guangxuan Xiao, Tianwei Yin, William~T Freeman, Fr{\'e}do Durand, and Song Han.
\newblock Fastcomposer: Tuning-free multi-subject image generation with localized attention.
\newblock \emph{IJCV}, 2024.

\bibitem[Xu et~al.(2022)Xu, Zhang, Zhang, and Tao]{xu2022vitpose}
Yufei Xu, Jing Zhang, Qiming Zhang, and Dacheng Tao.
\newblock Vitpose: Simple vision transformer baselines for human pose estimation.
\newblock In \emph{NeurIPS}, 2022.

\bibitem[Yang et~al.(2024)Yang, Ge, Li, Chen, Ge, Shan, and Chen]{yang2024seed}
Shuai Yang, Yuying Ge, Yang Li, Yukang Chen, Yixiao Ge, Ying Shan, and Yingcong Chen.
\newblock Seed-story: Multimodal long story generation with large language model.
\newblock \emph{arXiv preprint arXiv:2407.08683}, 2024.

\bibitem[Yang et~al.(2023)Yang, Wang, Gan, Li, Lin, Wu, Duan, Liu, Liu, Zeng, et~al.]{yang2023reco}
Zhengyuan Yang, Jianfeng Wang, Zhe Gan, Linjie Li, Kevin Lin, Chenfei Wu, Nan Duan, Zicheng Liu, Ce~Liu, Michael Zeng, et~al.
\newblock Reco: Region-controlled text-to-image generation.
\newblock In \emph{CVPR}, 2023.

\bibitem[Ye et~al.(2023)Ye, Zhang, Liu, Han, and Yang]{ye2023ip}
Hu~Ye, Jun Zhang, Sibo Liu, Xiao Han, and Wei Yang.
\newblock Ip-adapter: Text compatible image prompt adapter for text-to-image diffusion models.
\newblock \emph{arXiv preprint arXiv:2308.06721}, 2023.

\bibitem[Yue et~al.(2024)Yue, Wang, Sun, Ji, Chang, Zhang, et~al.]{yue2024exploring}
Zhongqi Yue, Jiankun Wang, Qianru Sun, Lei Ji, Eric~I Chang, Hanwang Zhang, et~al.
\newblock Exploring diffusion time-steps for unsupervised representation learning.
\newblock In \emph{ICLR}, 2024.

\bibitem[Zhang et~al.(2023)Zhang, Rao, and Agrawala]{zhang2023adding}
Lvmin Zhang, Anyi Rao, and Maneesh Agrawala.
\newblock Adding conditional control to text-to-image diffusion models.
\newblock In \emph{CVPR}, 2023.

\bibitem[Zhou et~al.(2025)Zhou, Ye, Liu, Ma, Wang, Qiu, Lin, Zhao, and Liang]{zhou2025explore}
Jiaming Zhou, Ke~Ye, Jiayi Liu, Teli Ma, Zifan Wang, Ronghe Qiu, Kun{-}Yu Lin, Zhilin Zhao, and Junwei Liang.
\newblock Exploring the limits of vision-language-action manipulations in cross-task generalization.
\newblock In \emph{The Thirty-Ninth Annual Conference on Neural Information Processing Systems}, 2025.

\bibitem[Zhou et~al.(2024)Zhou, Zhou, Cheng, Feng, and Hou]{zhou2024storydiffusion}
Yupeng Zhou, Daquan Zhou, Ming-Ming Cheng, Jiashi Feng, and Qibin Hou.
\newblock Storydiffusion: Consistent self-attention for long-range image and video generation.
\newblock In \emph{NeurIPS}, 2024.

\bibitem[Zhu et~al.(2025{\natexlab{a}})Zhu, Wang, Zhao, Zhang, and Zhang]{Zhu_2025_ICCV}
Beier Zhu, Ruoyu Wang, Tong Zhao, Hanwang Zhang, and Chi Zhang.
\newblock Distilling parallel gradients for fast ode solvers of diffusion models.
\newblock In \emph{ICCV}, 2025{\natexlab{a}}.

\bibitem[Zhu et~al.(2025{\natexlab{b}})Zhu, Wang, Zhu, Li, Li, Fang, Wang, Wang, and Zhang]{zhu2025dynamicmultimodalprototypelearning}
Xingyu Zhu, Shuo Wang, Beier Zhu, Miaoge Li, Yunfan Li, Junfeng Fang, Zhicai Wang, Dongsheng Wang, and Hanwang Zhang.
\newblock Dynamic multimodal prototype learning in vision-language models.
\newblock In \emph{ICCV}, 2025{\natexlab{b}}.

\end{thebibliography}
\bibliographystyle{iclr2026_conference}

\appendix
\clearpage
% \section{Appendix}
\noindent\textbf{APPENDIX.}
This appendix presents supplementary materials that extend the methodological details, experimental evaluations, and analytical discussions introduced in the main body of the paper. 

\section{Additional Implementation Details}\label{A-implementation_details}
\noindent\textbf{Extracting subject masks.}
We extract subject masks ($M_{\mathsf{id}}$ and $M_{n}$) by averaging the image-text cross-attention maps over all layers at the final denoising timestep, focusing specifically on subject-related tokens. Let $Q_\mathsf{img}$ denote the keys of image features and $K_\mathsf{sub}$ the keys of the subject-related tokens. For each cross-attention layer $l$, the unnormalized attention weights are computed as:
\begin{equation}
W_l = \frac{Q_\mathsf{img}K_\mathsf{sub}^\top}{\sqrt{d}},
\end{equation}
where $d$ is the feature dimension. We then average the attention weights across all $L$ layers:
\begin{equation}
W = \frac{1}{L} \sum_{l=1}^{L} W_l.
\end{equation}
We apply Otsu's thresholding~\cite{otsu1975threshold} to obtain the binary subject mask $M$:
\begin{equation}
M = \mathrm{Otsu}(W).
\end{equation}

\noindent\textbf{Selection of the most salient identity features.}
Our \oursIR~refines target images using the most salient reference features, which are determined by the OT plan. Specifically, the saliency score of the $i$-th identity feature is computed as:
 \begin{equation}
    s_i^\mathsf{OT} = \sum_{n=1}^{N} \left\langle T_n(i, :),\ 1 - C(i, :) \right\rangle,
\end{equation}
 The top-$\alpha$ index set $\mathcal{I}_i$ in Eq.~\ref{eq:cross_attn_w} contains indices with the $\alpha$ highest saliency scores.

\begin{figure}[h]
  \centering
  \includegraphics[width=1.0\linewidth]{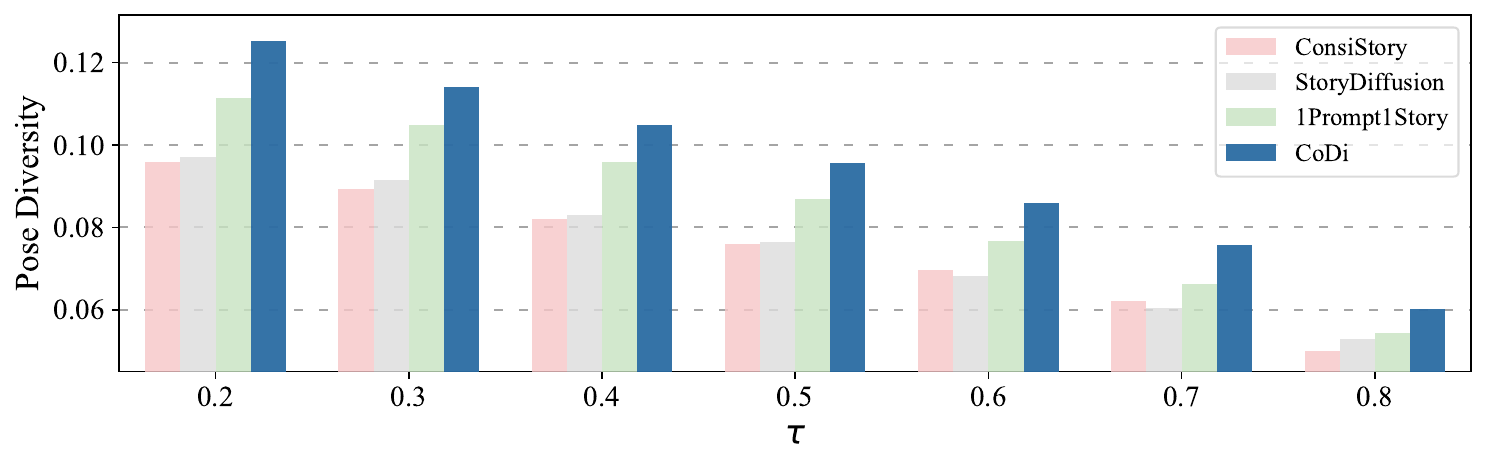}
  \caption{Pose diversity scores across different confidence thresholds $\tau$. Our \ours~ consistently outperforms other SCG methods by a clear margin under all $\tau$ settings.
  % and performs comparably to Vanilla SDXL~\cite{podell2023sdxl}. 
  }
  \label{A-fig:different_tau}
\end{figure}
\section{Additional Evaluation Details}\label{A-evaluation_details}
\noindent\textbf{Unified evaluation protocol.} We adopt a unified evaluation protocol across all metrics. Specifically,
for each target image set $k$ with $N$ generated images $\{\mathbf{x}_n\}_{n=1}^N$, we compute the average pairwise evaluation score as follows:
\begin{equation}
u_k = \frac{2}{N(N-1)} \sum_{n=1}^{N-1} \sum_{j=n+1}^{N} f(\mathbf{x}_n, \mathbf{x}_j),
\end{equation}
where $f(\cdot,\cdot)$ denotes the metric-specific similarity or distance function between two images, depending on the evaluation objective.
The final evaluation score is then obtained by averaging $u_k$ over all target image sets:\footnote{We slightly abuse the notations $K$, which here do not refer to keys in transformer.} 
\begin{equation}
    u = \frac{1}{K} \sum_{k=1}^{K} u_k.
\end{equation}
\noindent\textbf{Pose diversity score.}
We begin by extracting normalized 2D human keypoints and their confidence scores from each target image using ViTPose~\cite{xu2022vitpose}, a SoTA transformer-based model known for its high accuracy and robustness in human pose estimation. Each image $\mathbf{x}$ is represented by a set of $H$ keypoint locations $\mathbf{p}$ and their confidences $\bm{\beta}$.
\begin{equation}
\mathbf{p} = [(p_1^x,p_1^y)^\top,  \dots, (p_K^x,p_K^y)^\top]^\top \in \mathbb{R}^{H \times 2}, \quad \bm{\beta} = [\beta_1,\dots,\beta_K]^\top \in \mathbb{R}^K
\end{equation}
where each keypoint $\mathbf{p}_i = (p_i^x, p_i^y)$ is normalized by the image width and height and $\beta_i \in [0, 1]$ denotes its confidence score.
To ensure robustness, we discard keypoints with confidence scores below a threshold $\tau$. For a pair of target images $\mathbf{x}_i$ and $\mathbf{x}_j$, we retain only the indices of keypoints that are valid in both images.
We then perform Procrustes method~\cite{schonemann1966generalized} to remove global variations in translation, rotation, and scale by aligning $\mathbf{p}_i$ to $\mathbf{p}_j$. Specifically, 
we first compute the centroids of the keypoints which are denoted as  $\bm{\mu}_i$ and $\bm{\mu}_j$. 
We then center both keypoint sets by subtracting their respective centroids and normalize their $\ell_2$ norm:
\begin{equation}
\bar{\mathbf{p}}_i= \frac{\mathbf{p}_i - \bm{\mu}_i}{\|\mathbf{p}_i - \bm{\mu}_i\|_2}, \quad
\bar{\mathbf{p}}_j= \frac{\mathbf{p}_j - \bm{\mu}_j}{\|\mathbf{p}_j - \bm{\mu}_j\|_2}\end{equation}
Next, we compute the optimal rotation matrix using singular value decomposition ($\mathrm{SVD}$):
\begin{equation}
    \mathbf{U},\, \boldsymbol{\Sigma},\, \mathbf{V}^\top = \mathrm{SVD}(\bar{\mathbf{p}}_i^\top\bar{\mathbf{p}}_j), \quad \mathbf{R} = \mathbf{V}^\top \mathbf{U}^\top.
\end{equation}
The resulting $\mathbf{R}$ is an orthogonal rotation matrix that minimizes the Frobenius norm between the aligned keypoint sets, ensuring the best rigid alignment in the least-squares sense. The optimal scaling factor is given by:
\begin{equation}
    \gamma = \frac{\|\bar{\mathbf{p}}_j\|_2}{\|\bar{\mathbf{p}}_i\|_2} \cdot \mathrm{tr}(\boldsymbol{\Sigma}).
\end{equation}
The aligned keypoints are then obtained by applying the computed scale, rotation, and translation:
\begin{equation}
    \hat{\mathbf{p}}_i = \gamma \cdot \bar{\mathbf{p}}_i \mathbf{R} + \boldsymbol{\mu}_j.
\end{equation}
The pose diversity score between a pair of images $\mathbf{x}_i$ and $\mathbf{x}_j$ is computed as the average Euclidean distance between $\hat{\mathbf{p}}_i$ and $\mathbf{p}_j$.
To analyze pose diversity under different confidence thresholds $\tau$, we compare the pose diversity scores of various methods across a range of $\tau$ values. As shown in the Fig.~\ref{A-fig:different_tau}, our \ours~demonstrates a clear advantage over other SCG methods across all $\tau$ settings.
% our \ours~consistently outperforms other SCG methods under all $\tau$ settings.
% , and achieves performance comparable to Vanilla SDXL~\cite{podell2023sdxl}. 
We use $\tau = 0.7$ in our experiments to balance keypoint reliability and coverage.

\section{Limitations}
\begin{figure}[h]
  \centering
  \includegraphics[width=1.0\linewidth]{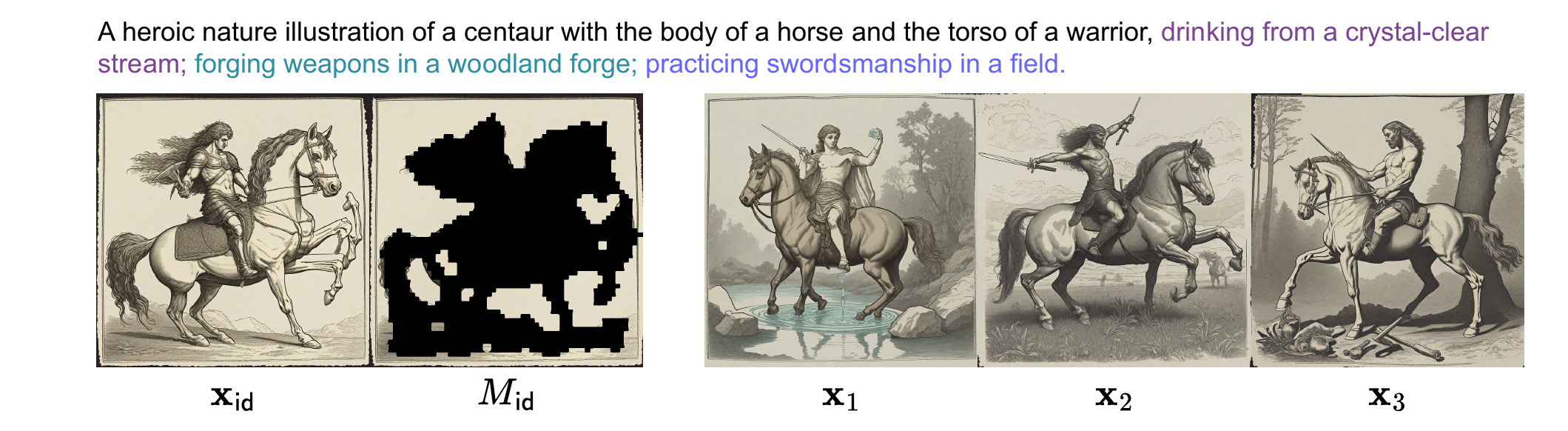}
  \caption{Limitations. Our method relies on the quality of cross-attention from the pre-trained diffusion model to accurately localize the subject.}
  \label{A-fig:limitation}
\end{figure}
Similar to prior subject-mask-based methods such as ConsiStory~\cite{tewel2024training}, our \ours~ framework relies on cross-attention scores to extract subject masks and estimate image token importance in the OT Plan. 
Occasionally, the pre-trained diffusion model assigns higher attention to background regions than to the subject, as shown in Fig.~\ref{A-fig:limitation}, which hinders the effective transport of identity features $\mathbf{X}_\mathsf{id}$ from the reference image $\mathbf{x}_\mathsf{id}$ to target image $\mathbf{x}_n$, resulting in subject inconsistency.
However,such failures are rare in practice (under 5\%) and can be solved by simply changing the seed.

\section{Inference Time and Memory Usage}
\begin{table}[h]
    \centering
    \caption{\textbf{Inference Time and Memory Usage}. We report the inference time (in seconds) and peak GPU memory usage on a single A6000 GPU for generating a set of five images from a prompt set with a resolution of $1024 \times 1024$. StoryDiffusion~\cite{zhou2024storydiffusion} is excluded due to excessive GPU memory consumption beyond the A6000’s limit.
}
    \label{tab:inference}
    \tabstyle{7pt}
    \begin{tabular}{l|cc}
    \toprule
    Method &  Inference time (s)    & GPU memory (GB) \\ 
    \midrule
    Vanilla SDXL~\cite{podell2023sdxl} & 77.67 & 35.58  \\
    1Prompt1Story~\cite{liu2025one} & 115.85 & 17.13  \\
    ConsiStory~\cite{tewel2024training} & 113.88 & 46.60 \\
        \rowcolor{lightCyan}
    \ours~{(ours)} & 154.89  & 45.20  \\
    \bottomrule
    \end{tabular}
\end{table}

We measure the inference time and memory usage of different SCG methods on a single A6000 GPU, as shown in Table~\ref{tab:inference}. We report the wall-clock time for generating a set of five images from a prompt set (since the baseline method ConsiStory~\cite{tewel2024training} performs cross-image attention across a batch of images) at a resolution of $1024 \times 1024$. Based on Table~\ref{tab:inference}, our method CoDi exhibits slightly higher inference time (154.89s) and comparable GPU memory usage (45.20GB) relative to ConsiStory. While 1Prompt1Story~\cite{liu2025one} is the most memory-efficient, it compromises subject consistency. Note that we exclude StoryDiffusion~\cite{zhou2024storydiffusion} due to excessive GPU memory usage beyond the A6000’s limit at $1024 \times 1024$ resolution (its original setting uses $768 \times 768$).

\begin{figure}[t]
  \centering
  % \vspace{-0.4cm}
  \includegraphics[width=1.0\linewidth]{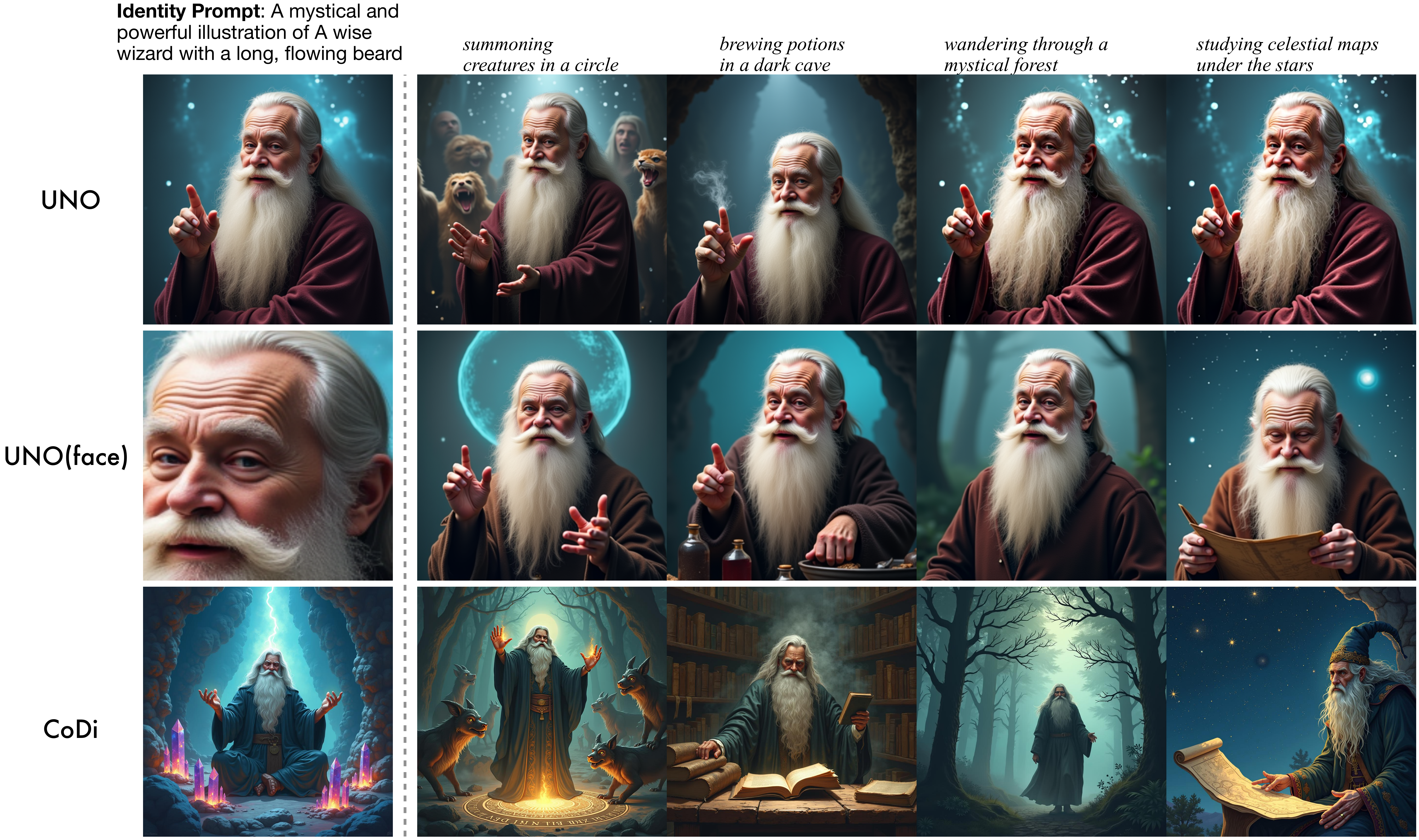}
  % \vspace{-0.7cm}
  \caption{\ours~shows superior results compared to SOTA training models.}
  \label{A-fig:compare_with_sota}
  % \vspace{-0.2cm}
\end{figure}

\section{Comparison with SOTA Training Method}

We compare \ours~with UNO~\cite{wu2025less}, as shown in Fig.~\ref{A-fig:compare_with_sota}. While UNO achieves subject consistency, its identical layout across outputs shows poor action prompt adherence and aesthetics. In contrast, \ours~demonstrates improved performance, highlighting the effectiveness of our training-free approach, which shows competitive results compared to SOTA company-level models.

\begin{figure}[t]
  \centering
  % \vspace{-1.5cm}
  \includegraphics[width=1.0\linewidth]{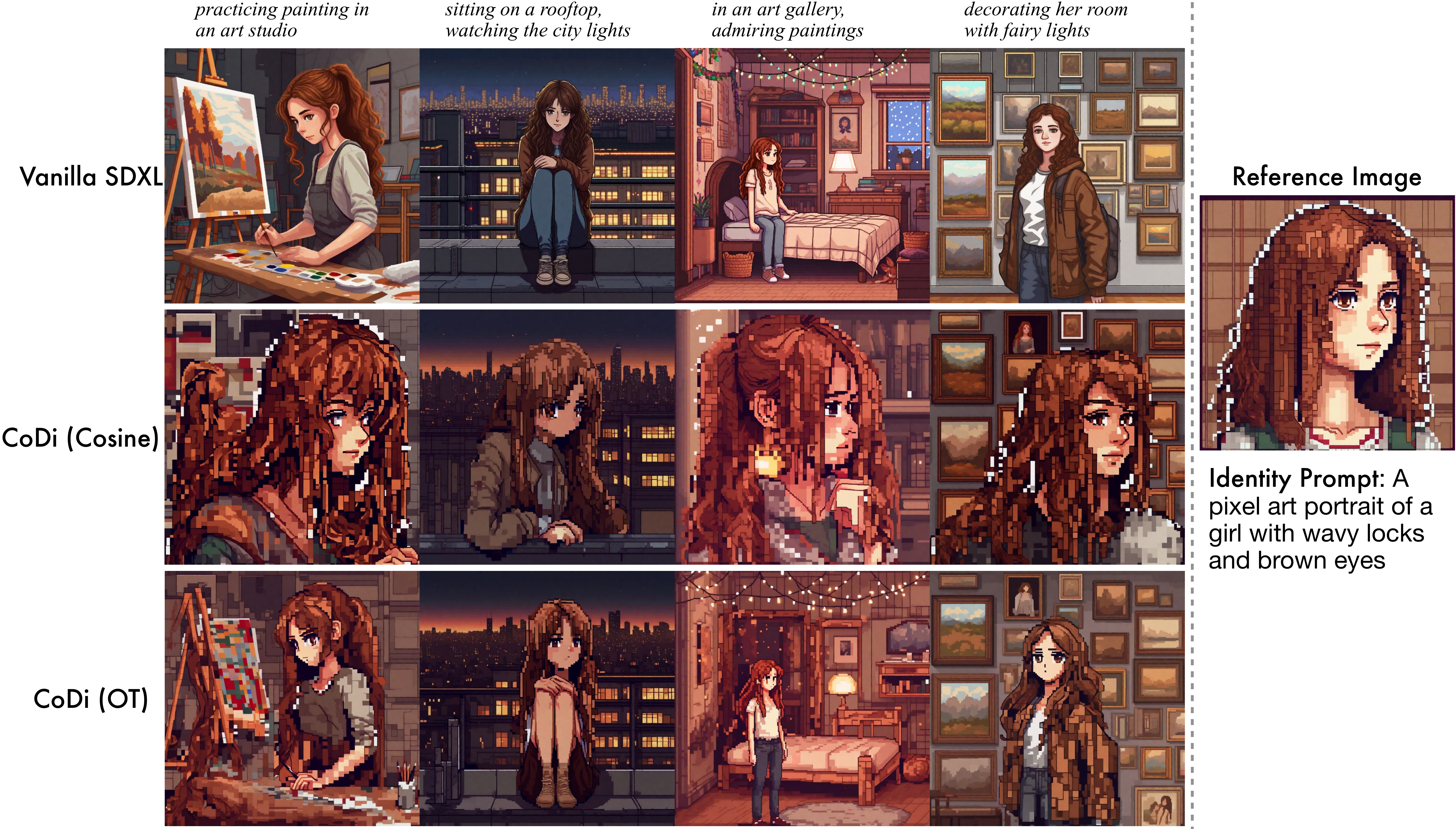}
  % \vspace{-0.6cm}
  \caption{CoDi (Cosine) fails to transport the corresponding features when there is a significant difference between the reference and target images. In contrast, CoDi (OT) achieves overall structural alignment, thereby preserving pose diversity.}
  \label{A-fig:OT_cosine}
\end{figure}

\section{The Effectiveness of Optimal Transport}

To assess the effectiveness of OT, we compare it with a simpler cosine similarity approach. CoDi (OT) achieves 0.0758, significantly outperforming CoDi (Cosine) at 0.0704 in pose diversity.
As shown in Fig.~\ref{A-fig:OT_cosine}, when there is a large feature difference between the reference and target images, especially across styles, CoDi (Cosine) tends to transport the mismatched features, failing to preserve the reference image’s pose. In contrast, CoDi (OT) effectively maintains the pose while enabling diverse variations, resulting in a failure to preserve the reference image’s pose. In contrast, CoDi (OT) effectively preserves the pose of the reference image while maintaining diverse pose variations.

\begin{figure}[h]
  \centering
  \includegraphics[width=1.0\linewidth]{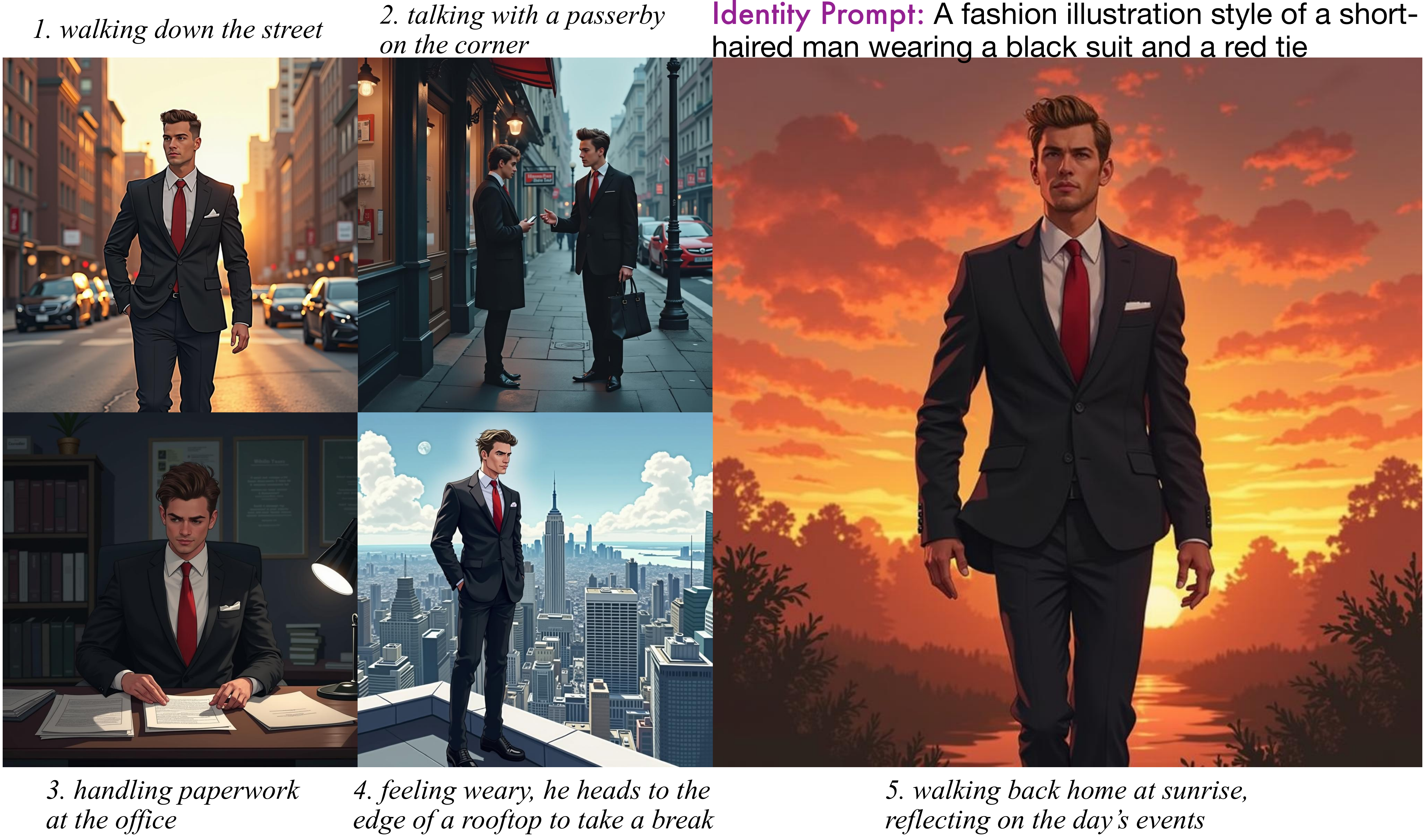}
  \caption{\ours~can maintain both subject and clothing consistency while preserving pose diversity by adding the clothing description in the prompt and adjusting the subject mask threshold to include clothing in the foreground mask.}
  \label{A-fig:long_story_generation_with_clothing_consistency}
\end{figure}
\section{Generalization to DiT-based Architectures}

\begin{figure}[t]
  \centering
  \includegraphics[width=1.0\linewidth]{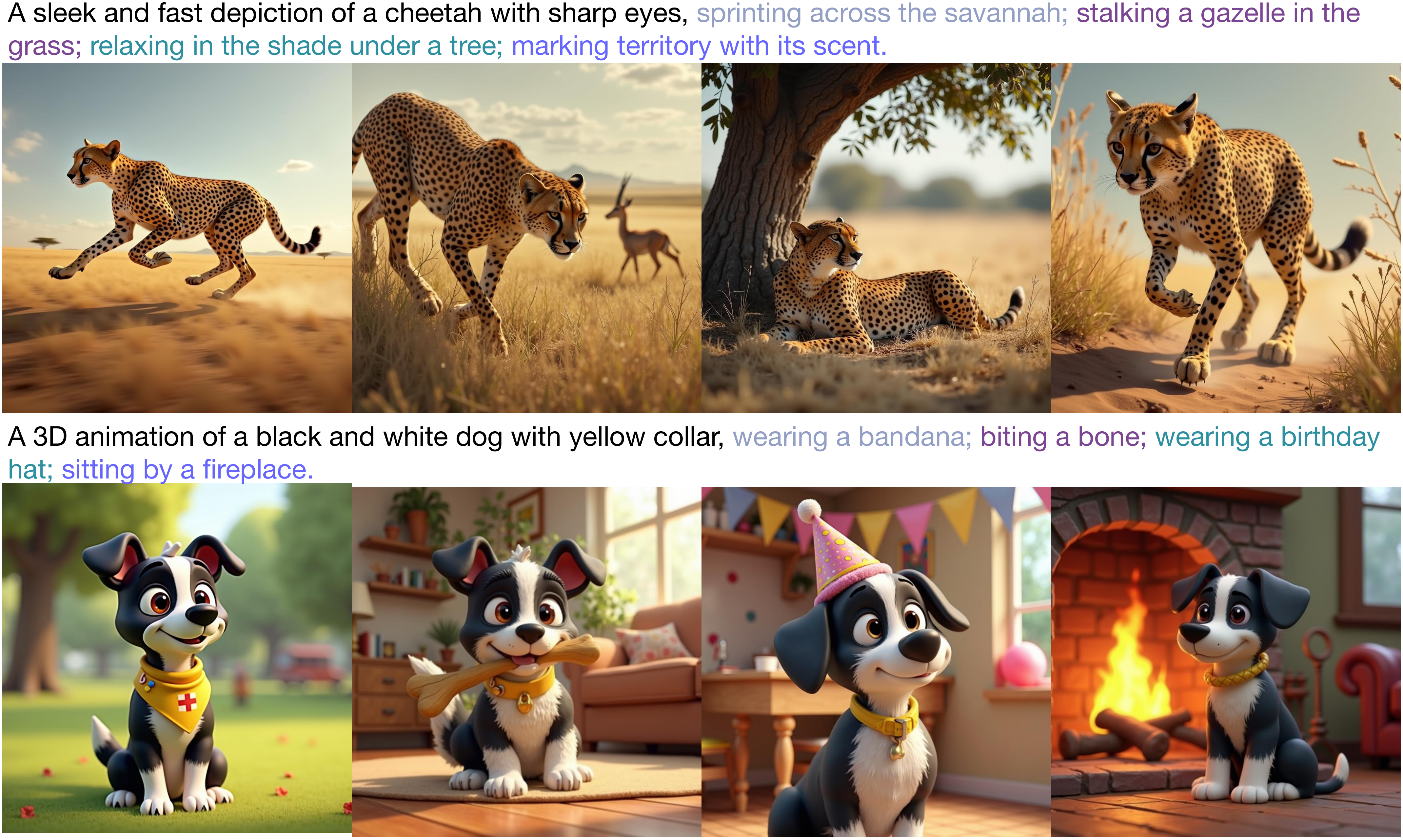}
  \caption{We have extended \ours~to the DiT-based architecture (Flux), with the generated results maintaining subject consistency while achieving diverse pose diversity.}
  \label{A-fig:flux_generated_result}
\end{figure}

We adapt \ours~to DiT-based models (Flux), with the generation results shown in Fig.~\ref{A-fig:flux_generated_result}, demonstrating richer details, subject consistency, and pose diversity.

\section{User Study Details} \label{user_study_details}

We conducted a user study comparing our method with state-of-the-art approaches, including ConsiStory, StoryDiffusion, and 1Prompt1Story. Thirty prompt sets, each containing four fixed-length prompts, were randomly sampled. Thirty-nine participants evaluated which method achieved the best overall image quality in terms of subject consistency, pose/layout diversity, and prompt fidelity.

\begin{figure}[t]
  \centering
  \includegraphics[width=1.0\linewidth]{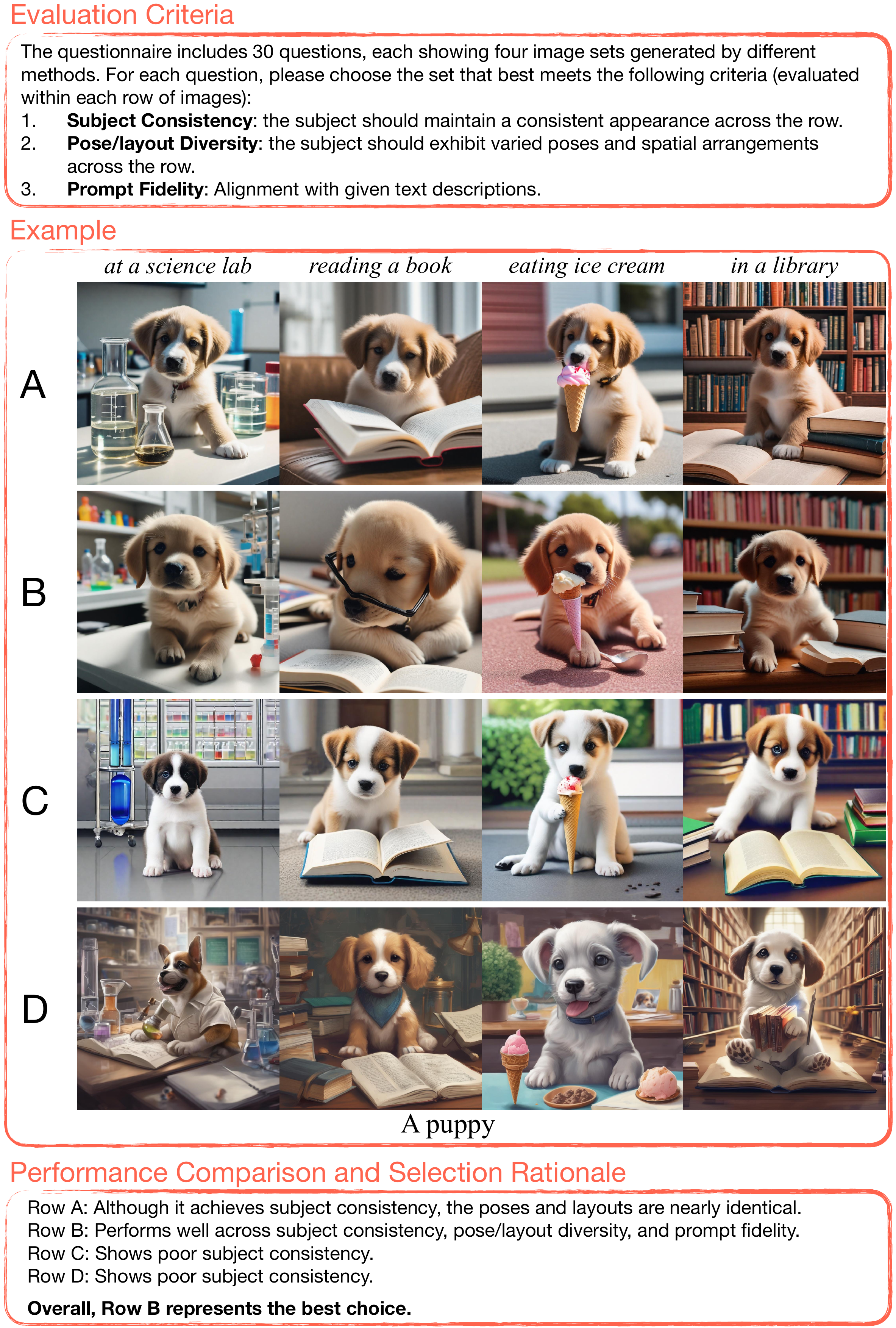}
  \caption{\textbf{Questionnaire of user study.} Evaluation Criteria outlines the standards for selecting image sets. Example illustrates a representative question for demonstration purposes. Performance Comparison and Selection Rationale demonstrates how the best choice is determined based on the example. These sections visually convey the evaluation criteria and guide participants’ selections.}
  \label{A-fig:user_study_guidance}
\end{figure}

Participants were instructed to select the set that best satisfied three evaluation criteria: subject consistency, pose/layout diversity, and prompt fidelity. Fig.~\ref{A-fig:user_study_guidance} illustrates these criteria at the start of the questionnaire. To facilitate informed selections, a representative example was provided, accompanied by a performance comparison and rationale, highlighting the reasoning behind choosing the optimal set.

\section{Additional Results}
\begin{figure}[t]
  \centering
  \includegraphics[width=0.92\linewidth]{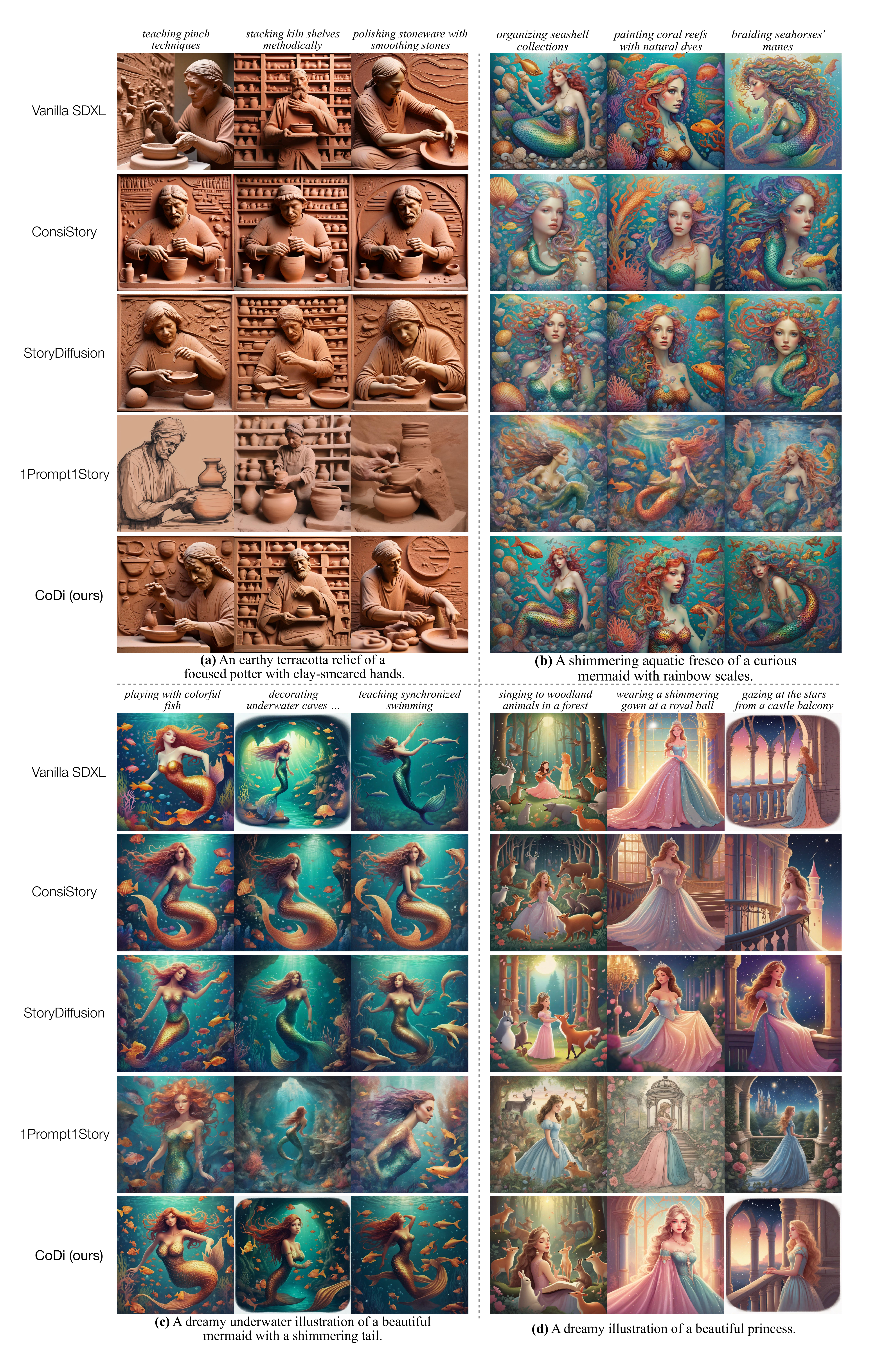}
  \caption{Additional qualitative comparisons. Our \ours~ achieves the best trade-off among subject consistency, pose diversity, and prompt fidelity.
}
  \label{A-fig:additonal_qualitative_comparison}
\end{figure}

\begin{figure}[t]
  \centering
  \includegraphics[width=1.0\linewidth]{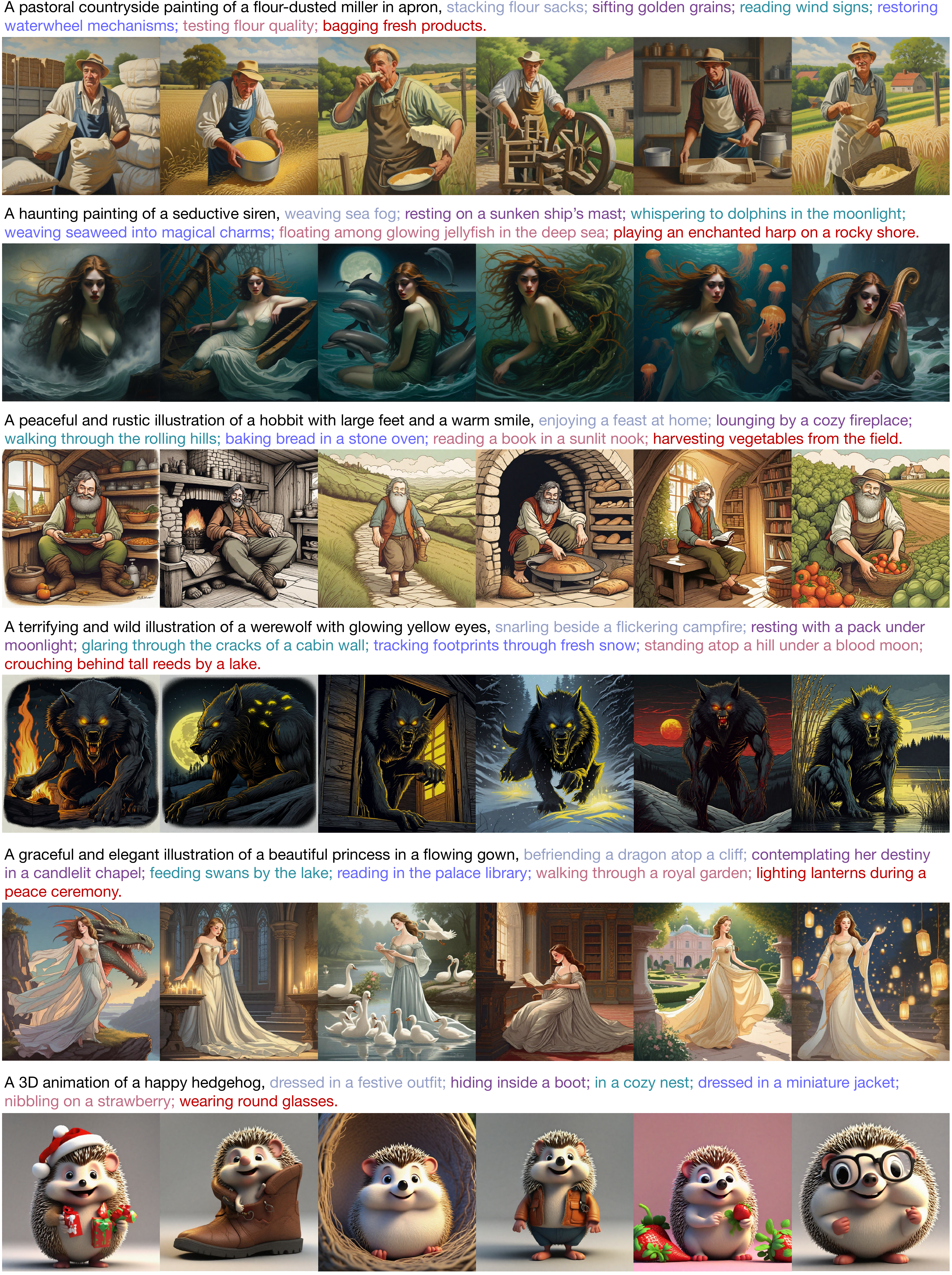}
  \caption{Additional qualitative results generated by our \ours~ demonstrate strong subject consistency and pose diversity.}
  \label{A-fig:additonal_qualitative_results}
\end{figure}

We present additional qualitative comparisons in Fig.~\ref{A-fig:additonal_qualitative_comparison}, along with more results generated by our \ours~ in Fig.~\ref{A-fig:additonal_qualitative_results}. 
These examples further demonstrate that our method achieves state-of-the-art performance in subject consistency, pose diversity, and prompt fidelity. In contrast, existing SCG methods remain limited, often excelling in only one or two of these aspects—typically at the expense of pose diversity or subject consistency.

\begin{figure}[t]
  \centering
  \includegraphics[width=1.0\linewidth]{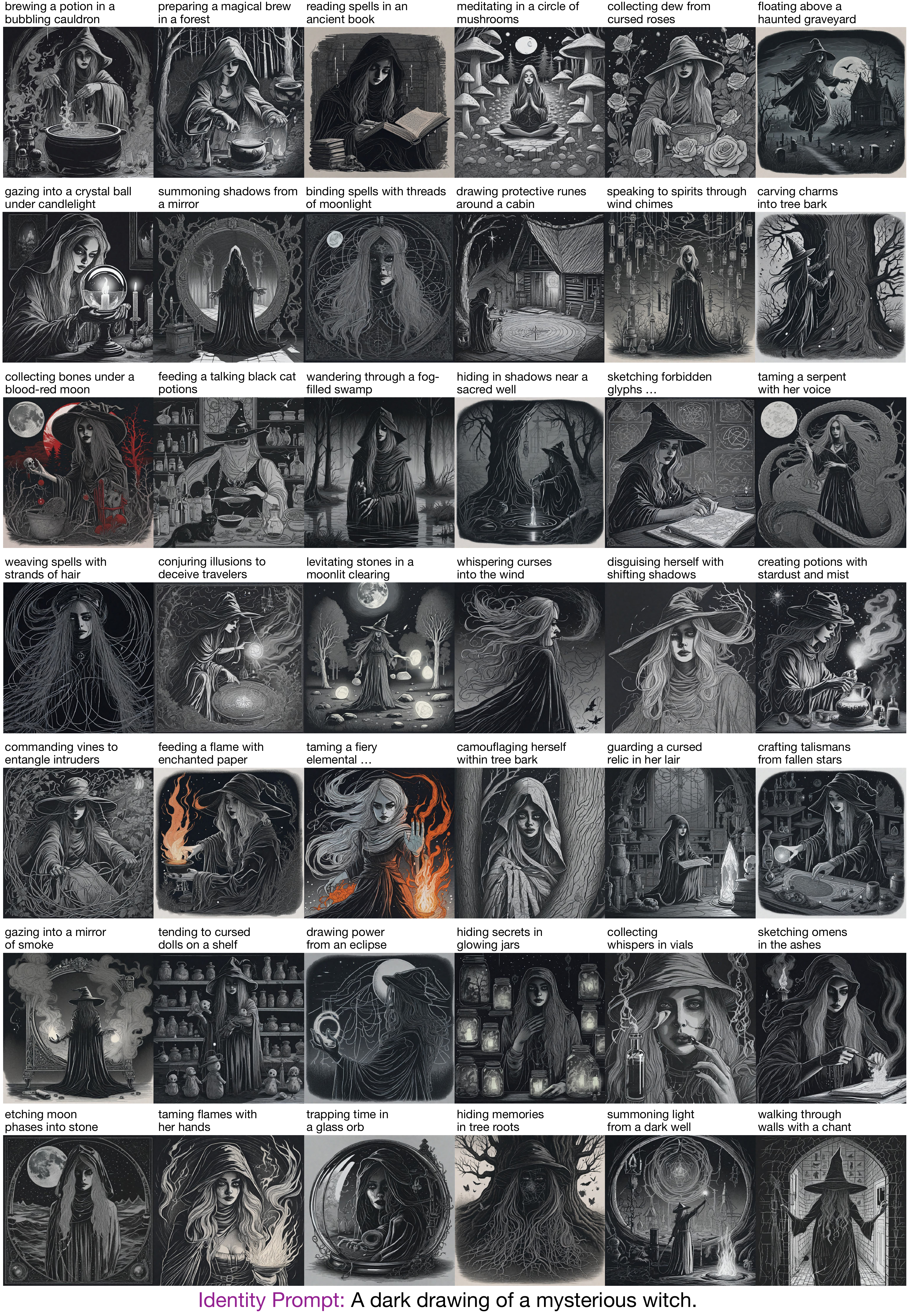}
  \caption{Long Story Generation. \ours~ supports extended visual storytelling by generating diverse scene compositions while consistently preserving subject identity throughout the sequence.}
  \label{A-fig:long_story}
\end{figure}

\noindent\textbf{Long story generation.}
As each target image $\mathbf{x}_n$ relies solely on reference image $\mathbf{x}_\mathsf{id}$ for subject identity, our \ours~ enables extended visual storytelling. As demonstrated in Fig.~\ref{A-fig:long_story}, it maintains subject consistency across diverse prompt semantics, supporting the generation of varied layouts and poses.
This makes \ours~ effective for long-form generation, where both prompt fidelity and visual diversity are essential.

\section{The Use of Large Language Models (LLMs)}
In our work, large language models (LLMs) were employed primarily for general writing assistance. Specifically, we used LLMs to refine sentence expressions, check for grammatical errors, and convert tables into \LaTeX{} format. 


% \input{A-sections/A-figures/Long_story_generation_with_clothing_consistency}
% \input{A-sections/A-figures/Different_style}


\end{document}